\PassOptionsToPackage{table}{xcolor}

\documentclass[table]{article}

\usepackage{placeins}
\usepackage{arxiv}
\usepackage{color,soul}
\usepackage[utf8]{inputenc} % allow utf-8 input
\usepackage[T1]{fontenc}    % use 8-bit T1 fonts
\usepackage{hyperref}       % hyperlinks
\usepackage{url}            % simple URL typesetting
\usepackage{booktabs}       % professional-quality tables
\usepackage{amsfonts}       % blackboard math symbols
\usepackage{nicefrac}       % compact symbols for 1/2, etc.
\usepackage{microtype}      % microtypography
\usepackage{lipsum}
\usepackage{graphicx}
\usepackage{tikz}
\usepackage{multicol}
\usepackage[utf8]{inputenc}
\usepackage{caption}
\usepackage{amsmath}
\usepackage{subcaption}
\usepackage{bigstrut}
\usepackage{adjustbox}
\usepackage{float}
\usepackage{array}
\usepackage{multirow}
\usepackage{colortbl}
\usepackage{authblk}

\usepackage{paralist}
\usepackage{nameref}
\usepackage{tabularx} % Required for full-width tables
\usepackage{booktabs} % For professional lines
\usepackage[
    style=authoryear, 
    maxcitenames=1,      % Force "et al." if there is more than 1 author
    mincitenames=1,      % Number of names to show before "et al."
    backend=biber, 
    uniquelist=false,  
    sorting=nyt
]{biblatex}
\newcolumntype{P}[1]{>{\centering\arraybackslash}p{#1}} 
\newcolumntype{C}[1]{>{\centering\arraybackslash}m{#1}}

\definecolor{mygray}{rgb}{.906, .902, .902}
\DeclareUnicodeCharacter{2061}{\leftexp{}}
\graphicspath{ {./images/} }

\title{CAHAL: Clinically Applicable resolution enHAncement for Low-resolution MRI scans
}

\author[1,*]{Sergio Morell-Ortega }
\author[1]{Ángela González-Cebrián}
\author[2]{Boris Mansencal}
\author[3]{Marien Gadea}
\author[4]{Roberto Vivo-Hernando}
\author[5]{Gregorio Rubio}
\author[6]{Fernando Aparici}
\author[7]{Maria de la Iglesia-Vaya}
\author[8]{Gwenaelle Catheline}
\author[2]{Pierrick Coupé}
\author[1]{José V. Manjón}

\affil[1]{Instituto de Aplicaciones de las Tecnologías de la Información y de las Comunicaciones Avanzadas (ITACA), Universitat Politècnica de València, Camino de Vera s/n, 46022, Valencia, Spain}
\affil[2]{CNRS, Univ. Bordeaux, Bordeaux INP, LABRI, UMR5800, in2brain, F-33400 Talence, France}
\affil[3]{Department of Psychobiology, Faculty of Psychology, Universitat de Valencia, Valencia, Spain}
\affil[4]{Instituto de Automática e Informática Industrial, Universitat Politècnica de València, Camino de Vera s/n, 46022, Valencia, Spain}
\affil[5]{Departamento de Matemática Aplicada, Universitat Politècnica de València, Camino de Vera s/n, 46022 Valencia, Spain}
\affil[6]{Área de Imagen Médica. Hospital Universitario y Politécnico La Fe. Valencia, Spain}
\affil[7]{Unidad Mixta de Imagen Biomédica FISABIO-CIPF. Fundación para el Fomento de la Investigación Sanitario y Biomédica de la Comunidad Valenciana - Valencia, Spain}
\affil[8]{Univ. Bordeaux, CNRS, UMR 5287, Institut de Neurosciences Cognitives et Intégratives d'Aquitaine, Bordeaux, France}

\affil[*]{Corresponding author email: \texttt{sermoor1@teleco.upv.es}}

\begin{document}
\vspace{-2.5em}
\maketitle

\begin{abstract}

Large-scale automated morphometric analysis of brain MRI is limited by the thick-slice, anisotropic acquisitions prevalent in routine clinical practice. Existing generative super-resolution (SR) methods produce visually compelling isotropic volumes but often introduce anatomical hallucinations, systematic volumetric overestimation, and structural distortions that compromise downstream quantitative analysis and diagnostic safety. To address this, we propose CAHAL (Clinically Applicable resolution enHAncement for Low-resolution MRI scans), a hallucination-robust, physics-informed resolution enhancement framework that operates directly in the patient's native acquisition space. CAHAL employs a deterministic bivariate Mixture of Experts (MoE) architecture routing each input through specialised residual 3D U-Net experts conditioned on both volumetric resolution and acquisition anisotropy, two independent descriptors of clinical MRI acquisition. Experts are optimised with a composite loss combining edge-penalised spatial reconstruction, Fourier-domain spectral coherence matching, and a segmentation-guided semantic consistency constraint. Training pairs are generated on-the-fly via physics-based degradation sampled from a large-scale real-world database, ensuring robust generalisation. Validated on T1-weighted and FLAIR sequences against generative baselines, CAHAL achieves state-of-the-art results, improving the best related methods in terms of accuracy and efficiency.
\end{abstract}


\newpage

\section{Introduction}
Every day, medical facilities worldwide acquire a large number of MRI scans to study the brain. In clinical environments, considerations such as patient comfort and limited acquisition times force scans to be acquired with only a sparse set of orientations, a limited number of slices, substantial slice thickness, and varying in-plane resolutions (ranging from 0.5 to 3 mm\textsuperscript{2}). These low-resolution conditions pose a significant obstacle to the use of automated morphometric analysis tools in demographic studies, which could eventually contribute to a deeper understanding of brain structure and functionality. For instance, well-known tools such as FreeSurfer \parencite{Fischl2012FreeSurfer}, FSL \parencite{Jenkinson2012FSL}, SPM \parencite{Penny2007StatisticalImages}, or AssemblyNet \parencite{Coupe2020} have been predominantly developed and validated on high-quality, isotropic 1 mm scans acquired in controlled research settings. Consequently, when there is a significant gap between the resolution of the training data and that of clinical acquisitions, the performance of these methods (particularly in tasks such as image registration and segmentation) degrades, ultimately compromising the accurate quantification of brain structures.

Super-resolution (SR) techniques have been proposed to bridge this gap. Early methods exploited self-similarity through patch-based reconstruction \parencite{Rousseau2013} and non-local upsampling \parencite{Manjon2010Non-localUpsampling}. With the transition to deep learning, volumetric models such as the multiscale 3D convolutional networks of \parencite{Pham2019} demonstrated the ability of hierarchical architectures to recover high-frequency anatomical details. Subsequent work diversified along several complementary directions. Self-supervised approaches, notably SMORE \parencite{Remedios2023Self-SupervisedGap}, circumvent the need for paired training data by learning from high-resolution intra-slice information, while implicit neural representations such as ArSSR \parencite{Wu2023} formulate upsampling as a continuous voxel function, enabling arbitrary-scale reconstruction. In parallel, domain-randomization strategies, pioneered by SynthSR \parencite{Iglesias2021JointContrast} and adopted by related frameworks including SynthSeg \parencite{Billot2023SynthSeg:Retraining}, SynthMorph \parencite{Hoffmann2022SynthMorph:Images}, Learn2Synth \parencite{Hu2024Learn2Synth:Segmentation}, and SuperSynth \parencite{Liu2025AImaging}, synthesize training images across arbitrary contrasts, orientations, and resolutions from anatomical label maps, substantially broadening the applicability of SR to heterogeneous clinical data. Domain-adaptation methods based on generative adversarial networks (GANs) \parencite{Peng2022DA-VSR:Images, Cong2024DDASR:Images} offer an alternative by aligning source images to a target domain. More recently, generative architectures, from GANs \parencite{ZhuUnpairedNetworks, WangDISGAN:Cleaning} and normalizing flows \parencite{Beizaee2025HarmonizingHarmonization} to diffusion models \parencite{Li2022SRDiff:Models}, have been applied to model complex image distributions, with frameworks such as MoEDiff-SR \parencite{Wang2025} integrating probabilistic mixture-of-experts routing for region-specific denoising.

Despite their capacity to produce visually compelling results, generative SR methods face three fundamental limitations that hinder clinical deployment. First, as demonstrated by \parencite{Cohen2018DistributionTranslation}, distribution-matching losses inherently risk introducing hallucinations, synthetic features absent from the true underlying anatomy. This risk is especially acute in domain-randomization approaches that rely on anatomical label maps: pathologies not represented in those labels, such as lesions, may be replaced by hallucinated healthy tissue, effectively erasing clinically relevant information. The work of \parencite{Riedel2026DeepDiseases} recently confirmed this phenomenon empirically, showing that generative models such as SynthSR systematically overestimate regional and global brain volumes, thereby distorting the atrophy patterns required for accurate automated disease prediction. Second, these methods introduce volumetric bias that directly undermines morphometric analysis, the primary use case for which SR is deployed in clinical neuroimaging. Third, an often-overlooked vulnerability arises at the pipeline integration level. Many automated preprocessing workflows register data to a standard template space, such as MNI152 \parencite{Fonov2009UnbiasedAdulthood, Fonov2011UnbiasedStudies}, prior to segmentation. Applying SR in the patient's native space before this step can improve registration quality. However, if the enhancement method alters anatomical boundaries, these errors propagate through every subsequent stage, from skull-stripping to tissue classification, ultimately compromising the validity of the final volumetric measurements.

In this paper, we adopt a deterministic, physically constrained approach to resolution enhancement that prioritises anatomical fidelity. Our framework is grounded in two complementary sources of real-world data drawn from the volBrain platform\footnote{Access at \href{https://volbrain.net/home}{https://volbrain.net/home}} \parencite{Manjon2016} and a regional clinical biobank. First, we leverage a large curated pool of high-resolution MRI scans ($\approx4,000$) sourced from volBrain, which has processed over 800,000 scans from diverse worldwide populations. This pool spans the full human lifespan and encompasses a wide range of anatomies, demographic backgrounds, and pathological conditions, including healthy controls, neurodegenerative cases, and white matter lesion burden varying from none to severe. Unlike label-to-image synthesis approaches such as SynthSR or SuperSynth, which generate anatomy from segmentation masks and therefore cannot represent pathologies absent from those masks, our ground truth consists of real patient images that inherently better encode the full spectrum of structural variability. Second, we exploit the rich acquisition metadata recorded by both volBrain and the regional biobank to model the empirical distribution of clinical MRI protocols. This enables the construction of a physically accurate degradation model that reflects the true heterogeneity of slice thicknesses and anisotropy profiles encountered in routine clinical practice, rather than relying on simplified or randomly sampled augmentations. Together, these two data sources allow the proposed framework to preserve pathological details present in the source anatomy and to generalise robustly to unseen clinical sites.

Motivated by these challenges, we propose a novel resolution MRI enhancement framework that operates directly in the patient’s native space. Although the present study demonstrates its efficacy on T1-weighted and FLAIR images, the methodology is highly adaptable and can be straightforwardly extended to other MRI modalities. We simulate realistic synthetic low-resolution data by degrading a curated pool of high-fidelity ground truth volumes, comprising research-grade scans. This work significantly expands upon the developments introduced in our earlier REMIX framework \parencite{Morell-Ortega2024REMIX:Experts} with four key contributions: a new deep network architecture and semantic loss, an expanded Mixture of Experts, native space processing (instead of MNI) and an extensive validation. We name our new proposed approach as CAHAL (Clinically Applicable resolution enHAncement for Low-resolution MRI scans).

\section{Materials and methods}

\subsection{High-resolution training data}
\label{subsec:hr_data}

The proposed framework relies on two complementary sources of real-world data. First, a curated pool of unique high-resolution MRI scans serves as the reconstruction target for the supervised super-resolution task. Second, the acquisition metadata recorded across large-scale clinical repositories provides the empirical foundation for the degradation model described in Section~\ref{subsec:degradation}. This dual role distinguishes our approach from label-to-image synthesis strategies such as SynthSR or SuperSynth: because our ground truth consists of real patient images rather than anatomy generated from segmentation masks, the training distribution inherently encodes the full spectrum of structural variability, including pathologies that label-based approaches cannot represent.

To ensure robust generalisation across clinical scenarios, we curated multi-source T1-weighted (T1w) and FLAIR datasets spanning a broad demographic and clinical spectrum. The selected volumes encompass a wide range of ages, geographical origins, and biological sex, as well as both healthy and pathological anatomies with varying lesion burdens.

For T1w images we utilized two primary sources:
\begin{itemize}
\item \textbf{Human Connectome Project (HCP) dataset:} We selected 1,027 T1w volumes from the HCP 
database~\parencite{VanEssen2012}. These scans are characterized by high isotropic resolution ($0.7 \text{ mm}$ isotropic, voxel volume $\approx 0.343 \text{ mm}^3$) and high signal-to-noise ratio (SNR), making them ideal for learning anatomical details. However, they are demographically limited to healthy young adults (ages 22–37).

\item \textbf{volBrain HR T1w curated dataset:} To increase the variability of the training data we used a curated subset of T1w images (N=3,715) from the \textbf{volBrain} platform (\href{volbrain.net}{https://volbrain.net/}) dataset \parencite{Manjon2016}. \textbf{volBrain} platform has processed online more than 800,000 cases during the last 10 years, from those cases, 470,000 have been donated by the users for research purposes. A total of 264,280 unique cases were identified based on the computation of volume hashes. The final high-quality subset was then selected according 
to strict criteria requiring voxel volume $< 0.125$~mm$^3$ 
and high contrast-to-noise ratio (CNR). While demographic metadata was unavailable for approximately 30\% of the donated cases, the selected subset spans an age range from 0 to 100 years, covering pediatric, adult, and geriatric morphologies from a global population (see Table \ref{tab:geographic_distribution}).
\end{itemize}

For the FLAIR modality, high-resolution ground truth data was sourced exclusively from the volBrain platform.
\begin{itemize}

\item \textbf{volBrain HR FLAIR curated dataset:} We curated a large-scale dataset from the \texttt{DeepLesionBrain} \parencite{Kamraoui2022DeepLesionBrain:Segmentation} pipeline for Multiple Sclerosis (MS) hosted on \textbf{volBrain} platform. Similar to our T1w protocol, we selected 1,098 high-quality samples from the user's database, applying strict quality control criteria (voxel volume $< 0.125 \text{ mm}^3$ and high CNR) to ensure their suitability as high-fidelity ground truth. This pool includes a wide range of pathological burdens, from healthy white matter to severe Multiple Sclerosis (MS). Specifically, we included cases with Total Lesion Volumes (TLV) up to $160.8$ $cm^3$ and Lesion Burdens (TLB) reaching $31.8\%$, ensuring the framework learns to preserve pathological boundaries without introducing inpainting artefacts.

\end{itemize}

Demographic profiles and pathological statistics are summarised in Table~\ref{tab:demographics}. Lifespan coverage and geographic distribution are detailed in Figure~\ref{fig:demographics_histograms} and Table~\ref{tab:geographic_distribution} in the Appendix.

\begin{table}[htbp]
\centering
\caption{\textbf{Summary of training data demographics and pathological diversity.} Data represents the high-resolution ground truth (HR) curated pool. For volBrain datasets, sex percentages are calculated based on the available metadata subset.}
\label{tab:demographics}
\resizebox{\columnwidth}{!}{%
\begin{tabular}{lcccccl}
\toprule
\textbf{Dataset} & \textbf{N} & \textbf{Age Range (Mean) in years} & \textbf{Sex (\% F/M/U)\textsuperscript{a}} & \textbf{Geography} & \textbf{Pathological Status} \\
\midrule
HCP \parencite{VanEssen2012} & 1,027 & 23-37 (28.84) & 54.39 / 45.60 / 0 & North America & Healthy (Young Adult) \\
volBrain T1w & 3,715  & 0-100 (47.65)  \textsuperscript{b} & 39.7 / 26.8 / 31.5  & Global & Real-world Clinical (Mixed) \\
\multirow{3}{*}{volBrain FLAIR} & \multirow{3}{*}{1,975} & \multirow{3}{*}{0--100 (49.01)\textsuperscript{b}} & \multirow{3}{*}{25.6 / 14.3 / 60.1} & \multirow{3}{*}{Global} & MS / Healthy \\
 &  &  &  &  & TLV: 0.0--160.8 $cm^3$ (Mean: 14.0)\textsuperscript{c} \\
 &  &  &  &  & TLB: 0.0--31.8\% (Mean: 3.5)\textsuperscript{d} \\
\bottomrule
\multicolumn{6}{l}{\textsuperscript{a} \footnotesize{F: Female, M: Male, U: Unknown.}} \\
\multicolumn{6}{l}{\textsuperscript{b} \footnotesize{Estimated range based on available metadata to show lifespan coverage.}} \\
\multicolumn{6}{l}{\textsuperscript{c} \footnotesize{TLV: Total Lesion Volume ($cm^3$). Median: 7.33; Std: 17.74.}} \\
\multicolumn{6}{l}{\textsuperscript{d} \footnotesize{LB (Lesion Burden) is calculated as the lesion volume divided by the total WM volume (in percent). Median: 1.79; Std: 4.50.}} \\
\end{tabular}
}
\end{table}

\subsection{Problem formulation and task scope}
\label{subsec:formulation}

The goal of CAHAL is to restore any low-resolution brain MRI acquisition to an isotropic resolution of $1 \times 1 \times 1$~mm$^3$ in the patient's native space, prior to any registration or segmentation step. Formally, given a low-resolution input 
$I_{LR}$ acquired with voxel size $(r_x, r_y, r_z)$, the input is first resampled to $1$~mm$^3$ via trilinear interpolation, providing an initial estimate of the target geometry. The network $f_\theta$ then operates entirely within this resampled space, learning to enhance its quality, recovering high-frequency anatomical detail and correcting partial volume artefacts, such that the output $\hat{I}_{HR}$ closely approximates a scan natively acquired at isotropic $1$~mm$^3$ resolution. Enhancement is only applied when at least one voxel dimension exceeds $1$~mm,
acquisitions already at or below this threshold require no processing.

Performing enhancement in native space, rather than after registration to a standard template, is a deliberate safety decision. Standard neuroimaging pipelines register 
clinical data to a template such as MNI152~\parencite{Fonov2009UnbiasedAdulthood, 
Fonov2011UnbiasedStudies} at $1$~mm$^3$ isotropic prior to segmentation. If a low-resolution scan is registered directly, the affine transformation compounds the existing resolution loss, degrading all subsequent processing steps. Applying SR in 
native space before registration avoids this cascade. However, it introduces a critical requirement: the enhancement method must not alter the underlying anatomy. Any hallucinated or displaced boundary will propagate through skull-stripping, tissue classification, and volumetric measurement, rendering all downstream outputs 
unreliable. This constraint motivates the deterministic, residual-based design described in Section~\ref{subsec:architecture}.

\subsection{Physics-informed degradation model}
\label{subsec:degradation}

\subsubsection{Ground truth preprocessing}
All high-resolution reference volumes were denoised prior to resampling using a spatially adaptive non-local means filter \parencite{Manjon2010AdaptiveLevels}, ensuring that the network learns to reconstruct anatomical structure rather than high-frequency noise. Next, the volumes were resampled to an isotropic resolution of $1 \times 1 \times 1$~mm$^3$ using cubic B-spline interpolation. This standardised volume serves as the ground truth $I_{GT}$ for training. Sourcing ground truth from ultra-high-resolution acquisitions (voxel volume $< 0.125$~mm$^3$) ensures that the resampled $1$~mm$^3$ target remains free of partial volume artefacts.

\subsubsection{Empirical acquisition analysis}

\begin{figure}[h!]
    \centering
    \includegraphics[width=\textwidth]{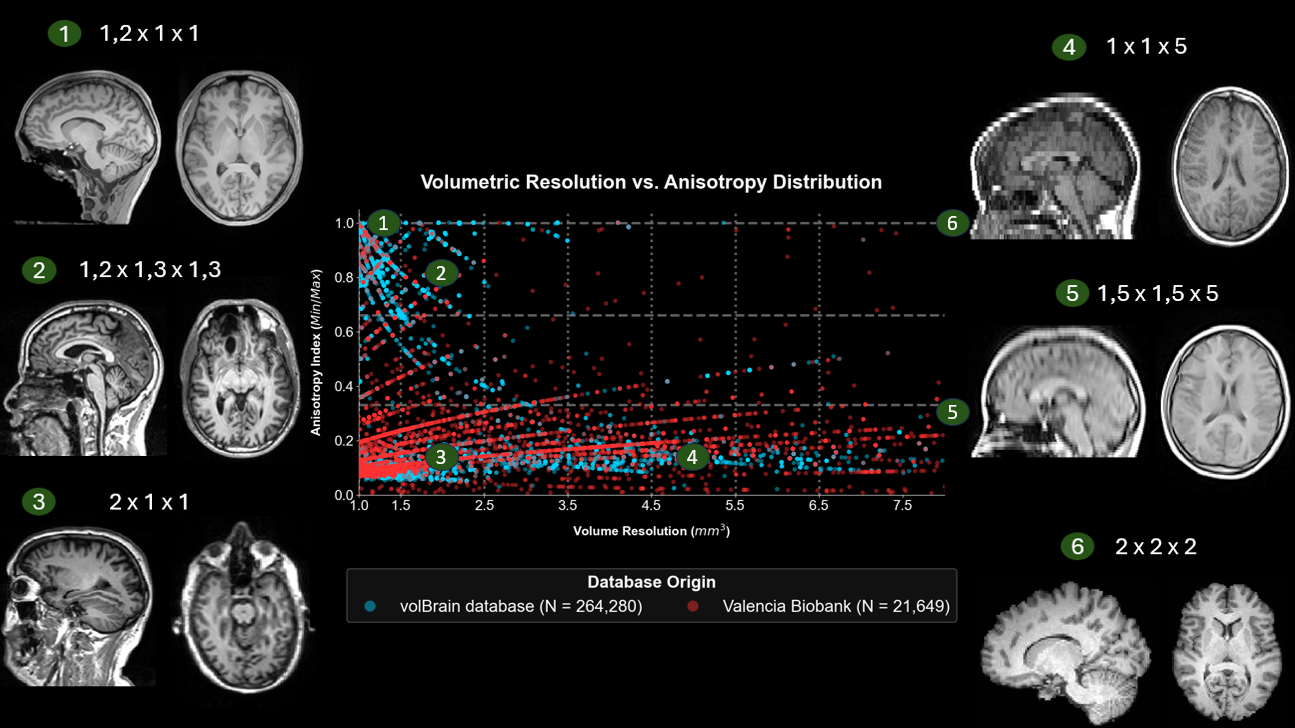}
    \caption{Left/right areas are some examples of real case resolutions and appearance. In the centre, the distribution of volume and anisotropy of our databases.}
    \label{fig:distribution_resolutions}
\end{figure}

To simulate realistic low-resolution inputs, acquisition parameters were sampled from the empirical distribution of the volBrain (N\,=\,264,280) and Valencia Biobank (N\,=\,24,694) databases (Figure~\ref{fig:distribution_resolutions} and Table~\ref{tab:resolution_dist}). The Valencia Biobank is a Spanish regional initiative that aggregates medical imaging data from multiple local hospitals for retrospective research. Together, these repositories reflect the true heterogeneity of clinical and research acquisition parameters across global populations.

This analysis revealed two findings that directly constrain the degradation model. First, over 99.9\% of scans maintain resolution consistency across at least two spatial dimensions, confirming that fully heterogeneous voxel geometries (where $r_x \neq r_y \neq r_z$) are negligible in practice. Second, while the volBrain research platform is dominated by isotropic acquisitions (58.6\%), the routine-clinical Biobank is almost fully anisotropic (98.5\%), demonstrating that isotropic degradation models are insufficient for real-world clinical deployment.

A further observation motivates the joint parameterisation of volume and anisotropy. A scan acquired at $1 \times 1 \times 5$~mm$^3$ and one at $1.7$~mm isotropic share a similar voxel volume of approximately $5$~mm$^3$, yet present fundamentally different restoration challenges: the former requires directional through-plane super-resolution while the latter requires isotropic deblurring. As illustrated in Figure~\ref{fig:distribution_resolutions}, these two protocols are indistinguishable along the volumetric axis but are widely separated along the anisotropy axis, demonstrating that volumetric resolution alone is an insufficient descriptor of the degradation physics.

\begin{table}[H]
\centering
\caption{Global demographics and distribution of resolution patterns across research and clinical databases. Resolution pattern tolerance is 0.1 mm. Sex is reported as Male/Female/Unknown (F/M/U).}
\label{tab:resolution_dist}
\resizebox{0.85\columnwidth}{!}{%
\begin{tabular}{@{}llrrrr@{}}
\midrule
 & & \multicolumn{2}{c}{\textbf{volBrain (Global)}} & \multicolumn{2}{c}{\textbf{Valencia Biobank (Clinical)}} \\ 
\midrule
\multicolumn{2}{@{}l}{\textbf{Demographics}} & & & & \\ \midrule
\multicolumn{2}{@{}l}{Total Scans} & \multicolumn{2}{c}{264,280} & \multicolumn{2}{c}{24,694} \\
\multicolumn{2}{@{}l}{Sex (\% F/M/U)} & \multicolumn{2}{c}{27.2 / 24.7 / 48.2} & \multicolumn{2}{c}{53.1 / 46.9 / 0} \\
\multicolumn{2}{@{}l}{Age Range (Mean) in years} & \multicolumn{2}{c}{0-100 (44.96)} & \multicolumn{2}{c}{0-100 (50.28)} \\ 
\midrule
\midrule
\textbf{Resolution Pattern} & \textbf{Likely Orientation} & \textbf{Count} & \textbf{\%} & \textbf{Count} & \textbf{\%} \\ \midrule
$X = Y = Z$ & Isotropic (3D) & 154,831 & 58.6 & 323 & 1.5 \\
$X = Y \neq Z$ & Axial (2D) & 63,786 & 24.1 & 21,326 & 98.5 \\
$Y = Z \neq X$ & Sagittal (2D) & 40,278 & 15.2 & - & - \\
$X = Z \neq Y$ & Coronal (2D) & 5,370 & 2.0 & - & - \\
$X \neq Y \neq Z$ & Heterogeneous & 15 & < 0.1 & - & - \\ \bottomrule
\end{tabular}%
}
\end{table}

\subsubsection{Degradation pipeline}

Low-resolution inputs are synthesised from $I_{GT}$ via a physics degradation function $\mathcal{D}$. Resolution parameters $(r_x, r_y, r_z)$ are sampled from the clinical distribution and clipped to a lower bound of $1.0$~mm:
$$\hat{r}_i = \max(r_i, 1.0\text{ mm}) \quad \forall i \in \{x, y, z\}$$

Rather than applying standard bicubic downsampling, which interpolates intensity at discrete points, we employ Adaptive Average Pooling (PyTorch \texttt{interpolate(mode='area')}~\parencite{Paszke2019PyTorch:Library}). This operation accurately models the MRI signal integration process, in which the recorded intensity of each voxel represents the average signal over its spatial volume:
$$I_{LR}(v) = \frac{1}{|\Omega_v|} \int_{\Omega_v} I_{GT}(x)\, dx$$
where $\Omega_v \subset \mathbb{R}^3$ denotes the spatial region spanned by voxel $v$. This formulation reproduces the partial volume effect and signal averaging inherent in thick-slice acquisitions. The degraded volume is subsequently upsampled back to $1$~mm$^3$ via trilinear interpolation, providing the network's initialisation.

Three constraints derived from the empirical analysis in Table \ref{tab:resolution_dist} are enforced by the parametric generator $\mathcal{G}(V, A)$, which produces resolution triplets $\mathbf{r} = (r_x, r_y, r_z)$ for a target volumetric resolution $V$ and anisotropy coefficient $A = r_{min}/r_{max}$:

\begin{itemize}
    \item \textbf{In-plane consistency.} Since fully heterogeneous resolutions account for fewer than 0.01\% of clinical scans, applying independent random blurring to all three axes, a common practice in the literature, is physically unrealistic. The generator therefore enforces that at least two dimensions share the same resolution, reflecting the fixed in-plane pixel size determined by the scanner's field of view and acquisition matrix.

    \item \textbf{Anisotropy handling.} The sharp contrast between the research-centric volBrain data and the clinical Biobank data highlights the necessity of modeling anisotropy. While isotropic training might suffice for research datasets (volBrain: $58.6\%$ isotropic), it fails for clinical populations where $98.5\%$ of scans are anisotropic (Biobank). For high-anisotropy samples, the generator fixes the in-plane resolution to $1$~mm (consistent with the dominant $X = Y \neq Z$ pattern) and solves for the required slice thickness as $r_z \approx \sqrt[3]{V}$, ensuring that the target volumetric resolution is matched while respecting the physical constraints of slice-selective acquisition.

    \item \textbf{Orientation permutation.} To prevent the network from overfitting to a specific acquisition plane
(e.g., learning that degradation only occurs along the Z-axis, typical of Axial scans), the generated resolution triplet is randomly permuted across $(r_x, r_y, r_z)$. This simulates the full range of axial, sagittal, and coronal acquisition orientations encountered in practice.
\end{itemize}

To constrain the problem space and facilitate network convergence, we standardised the orientation of all input volumes before processing. Medical imaging data can be stored with varying voxel orderings (e.g., LAS, PIL, RPS) depending on the scanner and acquisition protocol. We adopted a lossless Canonical Reorientation strategy, similar to the approach described in SuperSynth \parencite{Liu2025AImaging}. We reordered the data array in memory to conform to the standard Right-Handed-Coordinate System (RAS). This process involves strictly orthogonal operations, axis permutations (dimension swaps) and axis flips (reversing voxel order), based on the image header's direction cosine matrix. Crucially, this step is interpolation-free. It guarantees that the anatomical "up," "front," and "right" directions are consistently aligned with the network's logical axes ($x, y, z$) without altering the intensity values or spatial frequency content of the original data. This geometric standardization significantly reduces the manifold of variations the network must model, ensuring that the model focuses its capacity on resolving structural details rather than learning rotational invariance.

Training is performed on an infinite stream of synthetic pairs generated on-the-fly on GPU. At each iteration, a fresh resolution triplet is sampled from $\mathcal{D}$ and applied to a randomly selected ground truth volume, ensuring the network almost never encounters the same degraded realisation twice. This strategy effectively generates a virtually unlimited number of unique training pairs from the finite ground truth pool, preventing memorisation and providing robust coverage of the clinical degradation manifold throughout training.

% ---------------------------------------------------------------
\subsection{Bivariate Mixture of Experts}
\label{subsec:architecture}

Initially, we explored a unified modeling approach, training a single network on the full spectrum of simulated acquisitions. However, we hypothesized that such a "generalist" strategy (exemplified by frameworks such as SynthSR or SuperSynth) would suffer from capacity dilution when tasked with resolving the highly non-linear and heterogeneous degradation manifolds of clinical MRI. Relying on a single set of parameters to simultaneously master conflicting tasks (e.g., isotropic deblurring vs. highly anisotropic through-plane super-resolution) forces the network to find a "mean" solution that may be suboptimal for specific edge cases. This limitation is not merely theoretical: our ablation experiments (Section~\ref{sec:ablation}) confirm that a single-model baseline is consistently outperformed across all degradation regimes once the problem space is partitioned appropriately.

To address this, CAHAL adopts a Mixture of Experts (MoE) strategy,
which decomposes a complex modelling task into a set of more tractable sub-problems, each handled by a dedicated expert network. A gating mechanism then determines which expert to invoke based on the characteristics of the input. This philosophy has proven highly effective in large-scale AI systems, most notably in 
large language models such as Mixtral~\parencite{Jiang2024MixtralExperts} or even on the MRI super-resolution task \parencite{Wang2025}, and is 
well suited to the resolution heterogeneity encountered in real-world MRI, where the optimal reconstruction strategy varies substantially across acquisition protocols.

In our prior work REMIX~\parencite{Morell-Ortega2024REMIX:Experts}, we applied this principle to MRI super-resolution by partitioning the input space into seven clusters based on volumetric resolution, the product of the three spatial dimensions, ranging 
from $1.0$ to $1.5$~mm$^3$ up to $6.5$~mm$^3$ and beyond. While this strategy demonstrated the value of specialisation over a generalist model, volumetric resolution alone is an insufficient descriptor of the degradation physics, as established in 
Section~\ref{subsec:degradation}: two acquisitions with identical voxel volumes but different anisotropy present fundamentally different restoration challenges and require distinct reconstruction mappings.

CAHAL addresses this limitation by extending the clustering to a bivariate grid parameterised by volumetric resolution $V = r_x \cdot r_y \cdot r_z$ and anisotropy coefficient $A = r_{\min}/r_{\max} \in (0,\,1]$, where $A = 1$ denotes a perfectly isotropic acquisition and $A \to 0$ denotes extreme thick-slice anisotropy. Volumetric resolution is partitioned into seven bins: $V_1 \in [1.0,\,1.5)$,$V_2 \in [1.5,\,2.5)$, $V_3 \in [2.5,\,3.5)$,
$V_4 \in [3.5,\,4.5)$, $V_5 \in [4.5,\,5.5)$,
$V_6 \in [5.5,\,6.5)$, and $V_7 \in [6.5,\,+\infty)$~mm$^3$. Anisotropy is partitioned into three bins: $A_3 \in (0.66,\,1.0]$ (near-isotropic),
$A_2 \in (0.33,\,0.66]$ (moderate), and $A_1 \in (0,\,0.33]$ (highly anisotropic). Each cell in this grid defines a physically distinct degradation regime, ranging from mildly blurred near-isotropic scans to severely anisotropic thick-slice acquisitions, and is assigned a dedicated expert that concentrates its full capacity on that specific subset of restoration problems.

% \subsubsection{Expert network design}

For each expert network, we used a modified residual 3D U-Net~\parencite{Ronneberger2015} with four resolution levels and 64 initial filters, doubled at each downsampling step. Within each level, a convolutional layer is followed by instance normalisation and a LeakyReLU activation. A global residual connection links the input volume directly to the final layer. This design choice is both a technical and clinical decision: the residual formulation constrains the network to learn only the high-frequency correction required to restore the input, rather than synthesising an entirely new image. The final prediction is therefore anchored to the original native signal, making it less prone to hallucinating anatomy or inpainting pathological tissue. For volumes with large fields of view that exceed GPU memory, inputs are resampled to a fixed voxel count while preserving the aspect ratio.

% \subsubsection{Progressive curriculum training}

\begin{figure}[h!]
    \centering
    \includegraphics[width=\textwidth]{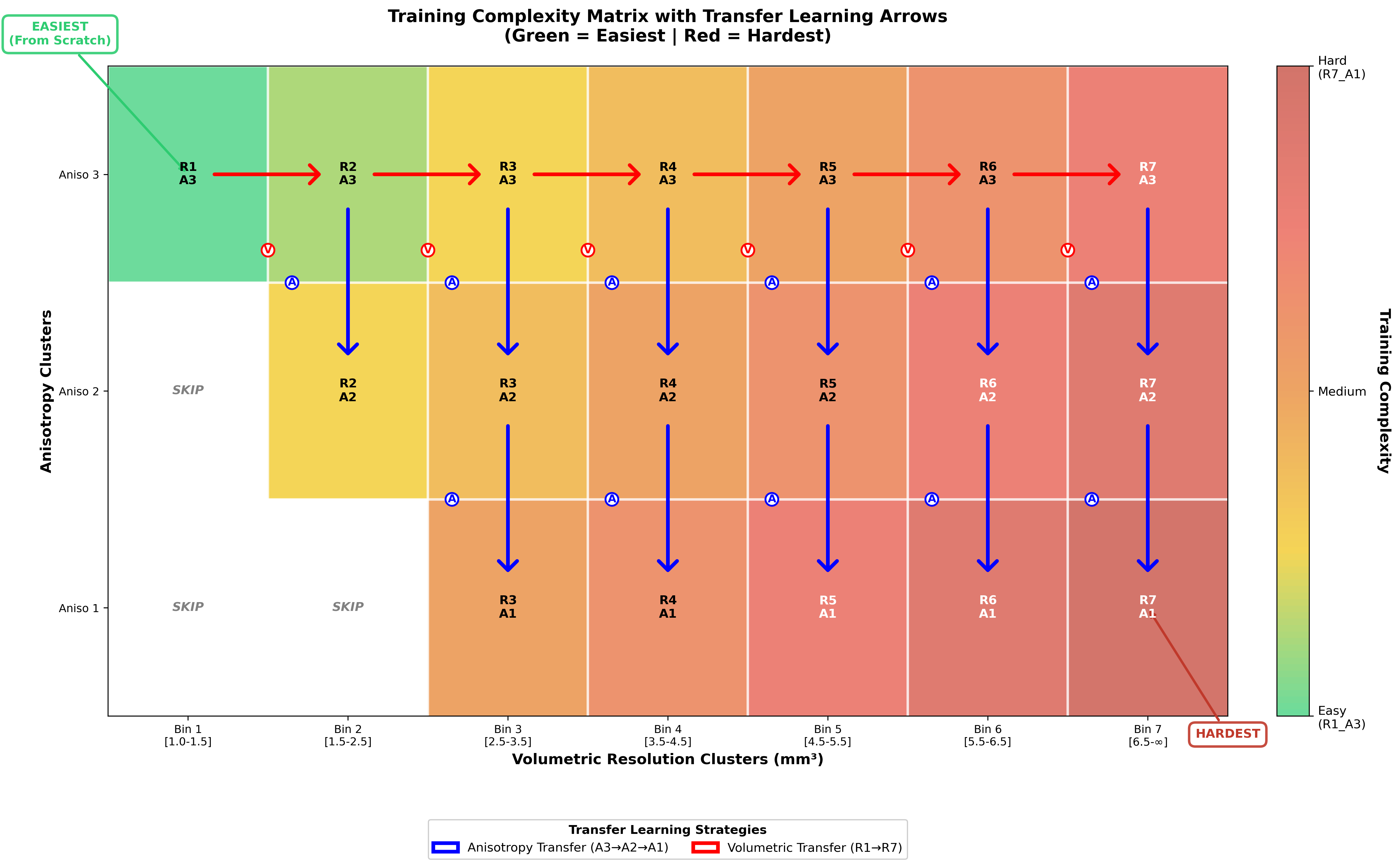}
    \caption{Progressive transfer learning strategy across the bivariate expert manifold. The matrix illustrates the training complexity for various degradation profiles, ranging from the easiest domain (green, trained from scratch) to the most severe (red). Rather than training all experts independently, we employ a cascading initialization approach. Horizontal red arrows denote volumetric transfer, sequentially passing learned weights across progressively lower spatial resolutions. Vertical blue arrows denote anisotropy transfer, adapting weights from nearly isotropic to highly anisotropic domains.}
    \label{fig:finetunning}
\end{figure}

Training follows a curriculum learning approach that navigates the bivariate expert grid from the simplest to the most complex degradation regime, as illustrated in Figure \ref{fig:finetunning}. The first expert, corresponding to the near-isotropic, lowest-volume cluster ($R1$-$A3$: $\sim 1$~mm$^3$ isotropic), is trained from scratch. Since training operates on an unbounded stream of synthetically generated pairs, convergence is monitored via the training loss, which serves as a reliable proxy for performance in this unbounded data regime. Its converged weights then initialise all adjacent experts via two transfer pathways: horizontal volumetric transfer passes weights across increasing voxel volumes at fixed anisotropy, while vertical anisotropy transfer adapts weights from near-isotropic to highly anisotropic regimes at fixed volume. This cascading initialisation provides an amortised training strategy: rather than optimising each expert independently from random weights, knowledge accumulated in simpler degradation regimes is progressively transferred to more complex ones.

Not all intersections in the bivariate grid correspond to physically realisable acquisitions. As established in Section~\ref{subsec:degradation}, the reconstruction target imposes a lower bound of $1$~mm on all spatial axes, since no enhancement is required along any dimension already at or below this threshold. This bound creates a hard geometric constraint on the joint distribution of $V$ and $A$. As the anisotropy is defined as $A = r_{min}/r_{max}$: a low anisotropy value (A1, corresponding to $A \in [0, 0.33]$) implies that the shortest axis is at most one 
third of the longest. If the shortest axis is constrained to a minimum of $1$~mm, the longest axis must be at least $3$~mm, and the voxel volume $V = r_x \cdot r_y \cdot r_z$ is therefore bounded below well above the $1.0$--$1.5$~mm$^3$ range of cluster V1. The same reasoning eliminates V1-A2 and V2-A1: in both cases, satisfying the anisotropy condition while respecting the $1$~mm lower bound forces the voxel volume beyond the cluster boundaries. This is not merely a theoretical argument: the empirical 
distributions of volBrain and the Valencia Biobank 
(Table~\ref{tab:resolution_dist}) contain no acquisitions 
in these subspaces, confirming that they are absent from 
real-world clinical data. The subspaces V1-A1, V1-A2, and V2-A1 are therefore excluded from the grid, concentrating computational resources exclusively on physically viable and empirically observed degradation patterns.

% \subsubsection{Training strategy}

\subsubsection{Deterministic gating}

Unlike probabilistic routing mechanisms that introduce stochastic expert selection at inference, CAHAL employs a fully deterministic gating strategy. Each input is routed to a single expert based exclusively on the volumetric resolution and anisotropy computed from the native image header metadata. This design guarantees predictable, reproducible outputs for any given acquisition and incurs zero additional computational overhead at inference: because exactly one expert is evaluated per input, run-time complexity remains $\mathcal{O}(1)$ regardless of the total number of experts in the grid, equivalent to running a single standard model.

\subsection{Training loss}

The training objective consists of two complementary components: a reconstruction loss that enforces voxel-level and spectral fidelity, and a segmentation-guided semantic loss. The core reconstruction Loss ($\mathcal{L}_{rec}$) of our network is formulated as a mixed loss that explicitly balances local spatial detail preservation with global frequency characteristics.

\begin{equation}
\mathcal{L}_{rec}(y, \hat{y}) = \mathcal{L}_{WMAE}(y, \hat{y}) + \lambda_{FFT} \mathcal{L}_{FFT}(y, \hat{y})
\label{eq:rec_loss}
\end{equation}

where $y$ is the expected output,  $\hat{y}$ is the predicted output. These two components address complementary aspects of 
restoration quality: spatial edge fidelity and global 
spectral coherence.
\begin{itemize}
    \item Weighted Mean Absolute Error (WMAE) ($\mathcal{L}_{WMAE}$):  In the spatial domain, we apply a voxel-wise weighting strategy to explicitly penalize errors along high-frequency structural edges. Rather than using raw intensity differences, we extract the structural edges from the ground truth image ($y$) by convolving it with a discrete 3D Laplacian kernel ($K_{Lap}$). To ensure numerical stability and prevent extreme outliers from dominating the loss gradients, the absolute magnitude of the Laplacian response is logarithmically compressed and normalized to a $[0, 1]$ range by dividing by its global peak value:

    % \begin{equation*}
    % W(y) = \frac{\ln(1 + \gamma |y * K_{Lap}|)}{\max(\ln(1 + \gamma |y * K_{Lap}|)) + \epsilon}
    % \end{equation*}

    \begin{equation*}
W(y) = \frac{\ln(1 + \gamma |y * K_{Lap}|)}{\max\limits_{x \in \Omega} \left( \ln(1 + \gamma |y * K_{Lap}|) \right) + \epsilon}
\end{equation*}

    where $*$ denotes the 3D convolution operation, $\gamma$ is a scaling factor (set to 1.0), and $\epsilon$ is a small constant ($10^{-6}$) to prevent zero-division. The final spatial loss incorporates this normalized edge map $W(y)$ as a penalty multiplier:
    \begin{equation*}
    \mathcal{L}_{WMAE}(y, \hat{y}) = \frac{1}{N} \sum_{i=1}^{N} \Big(1 + W(y_i)\Big) |y_i - \hat{y}_i|
    \end{equation*}
    
    where $N$ denotes the total number of voxels in the 3D volume. This formulation ensures that uniform, flat regions (e.g., homogeneous white matter) are optimized via standard Mean Absolute Error (where $(1 + W(y_i)) \approx 1$), while the sharpest anatomical boundaries receive up to double the penalization (where $(1 + W(y_i)) \approx 2$), effectively preventing the over-smoothing of complex cortical folds.
    
    \item Spectral Loss ($\mathcal{L}_{FFT}$): To explicitly recover the high-frequency spectral content lost during downsampling, we penalize differences in the 3D Fourier domain. To ensure numerical stability and prevent underflow when calculating large spectral magnitudes, this loss is strictly computed in full 32-bit precision (disabling automatic mixed precision during this step). The spectral term is scaled by an empirical factor of $\lambda_{FFT} = 2 \times 10^{-6}$ to balance its magnitude with the spatial WMAE, acting as a perceptual term that compels the network to restore global textural coherence. In a recent paper \parencite{Morell-Ortega2025RobustNetwork}, we demonstrated that FFT loss serves as a low memory demanding valid surrogate of perceptual loss in MRI contrast synthesis.  
    
\end{itemize}

To drive the reconstruction process beyond simple pixel-wise fidelity, we incorporated a segmentation-guided semantic loss, based on the Dice coefficient \parencite{Dice1945MeasuresSpecies}. This approach ensures that the enhanced images maintain clear anatomical boundaries and tissue identities consistent with downstream analysis tools.

Let $S(\cdot)$ represent a fixed, pre-trained segmentation network that maps an input volume to its multi-class tissue probability maps. The semantic loss evaluates the structural discrepancy between the segmentation of the ground truth and the segmentation of the super-resolved prediction:
\begin{equation}
\mathcal{L}_{seg}(y, \hat{y}) = 1 - \text{Dice}\Big(S(y), S(\hat{y})\Big)
\end{equation}

For $S(\cdot)$, we employed a Deep Pyramidal Network (DPN) \parencite{Morell-Ortega2025DeepCERES:MRI}, a lightweight variant of the 3D U-Net architecture. 

The DPN was selected specifically for its parameter efficiency and low memory footprint, which facilitates its direct integration into the training loop for online loss calculation without excessive computational overhead.

Since manual segmentation of our large-scale training set is infeasible, we used validated automated pipelines to generate dense semantic labels (proxy ground truth): 
\begin{itemize} 
\item \textbf{T1-weighted module:} Segmentation targets were generated using the \texttt{vol2brain} pipeline \parencite{Manjon2022Vol2Brain:Analysis}. This pipeline uses non-local patch-based priors to produce consistent segmentations across 7 tissue classes and is an evolution of \texttt{volbrain} pipeline \parencite{Manjon2016}. This technology is recognized for its robustness \parencite{Nss-Schmidt2016}. The T1-specific DPN was trained with the HR T1w and corresponding labels, achieving a mean Dice similarity coefficient (DSC) of 0.85 in native space. 
\item \textbf{FLAIR module:} Following the same protocol, we utilized the \texttt{DeepLesionBrain} pipeline \parencite{Kamraoui2022DeepLesionBrain:Segmentation} to generate multi-class maps (8 tissues including White Matter Lesions). The FLAIR-specific DPN was trained with the HR FLAIR and corresponding labels, achieving a mean DSC of 0.86 in native space. \end{itemize}

\begin{figure}[H]
    \centering
    \includegraphics[width=\textwidth]{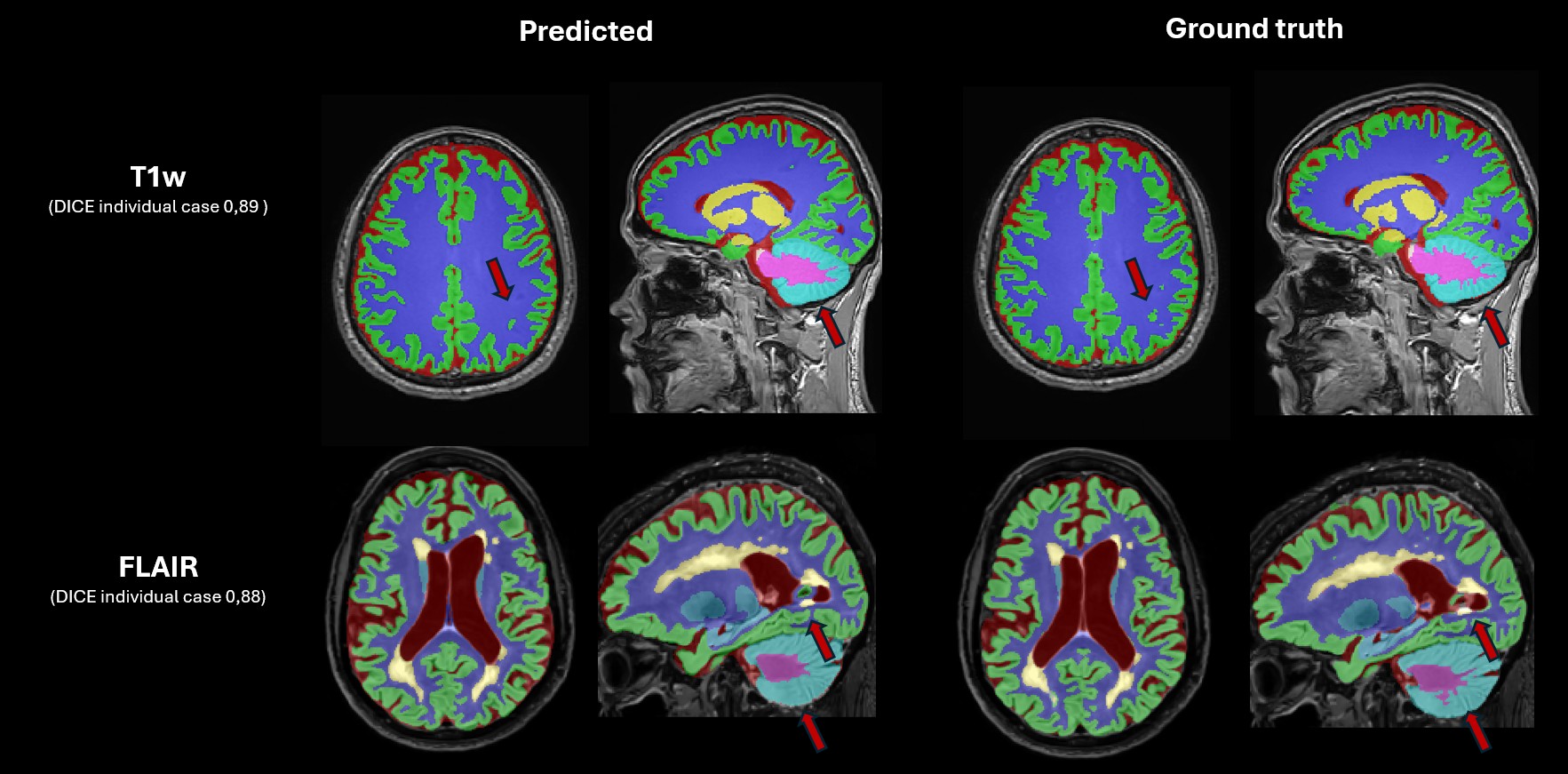}
    \caption{Visual comparison between the segmentations predicted by our DPN (left) and the proxy ground truth labels (right) for T1w (top) and FLAIR (bottom) acquisitions. Although some fine-grained morphological details are inevitably smoothed, the predicted masks accurately capture the macroscopic anatomy, serving as a highly reliable proxy for structural evaluation. Small differences are highlighted with red arrows.}
\label{fig:segmentations_generated}
\end{figure}

While these proxy labels may contain minor imperfections compared to manual delineation, they provide sufficiently strong semantic gradients to regularize the super-resolution task.  Furthermore, if the reconstructed image achieves a higher dice score, we can conclude that it has been better reconstructed. Figure \ref{fig:segmentations_generated} visualizes the correspondence between input images and the generated segmentation masks.

The final training objective is:

\begin{equation}
\mathcal{L}_{total} = \mathcal{L}_{WMAE}(y, \hat{y}) + \lambda_{FFT} \mathcal{L}_{FFT}(y, \hat{y}) + \lambda_{seg} \mathcal{L}_{seg}(y, \hat{y})
\end{equation}

where $\lambda_{seg} = 0.1$ is set to balance the magnitude of $\mathcal{L}_{seg}$ relative to the reconstruction loss $\mathcal{L}_{rec}$, defined in Equation \ref{eq:rec_loss}.

\section{Experiments and results}

The proposed framework was implemented in PyTorch 2.4 \parencite{Paszke2019PyTorch:Library}, with all training and evaluation procedures conducted on a single NVIDIA V100 GPU (\textbf{32 }GB VRAM). Across both the T1w and FLAIR experiments, we optimized the network weights using the Adam algorithm \parencite{Kingma2014}. To ensure robust convergence was achieved, we employed a dual sequential reduce on plateau scheduling strategy and an early stopping callback monitoring training loss (patience=25, early=75).

For T1 training, 4732 training cases were used, and for FLAIR, 1965 cases were used. We used no validation dataset because overfitting was practically impossible given our training-pair generation scheme.

For T1w and FLAIR evaluation, 10 high-quality subjects from volBrain (held out from training) serve as seeds. For each seed, synthetic acquisitions are generated across the 18 physically realisable clusters of the bivariate grid, excluding the three geometrically 
infeasible subspaces V1-A1, V1-A2, and V2-A1 as discussed above, with 3 independent realisations per cluster, yielding a total of 540 evaluation volumes ($10 \times 18 \times 3$). This structured evaluation set provides dense coverage of the low-resolution degradation manifold, ensuring that performance is assessed independently 
across every viable combination of resolution severity and anisotropy regime rather than on a single pooled degradation distribution.

In this section, we first present ablation experiments to analyze the contribution of the different components of the method to its final performance. We then compare our approach with related state-of-the-art methods, including an extensive set of validation experiments, where all statistical tests are done using the Mann-Whitney U two-sided test and Bonferroni corrections.  The reference baseline corresponds to the low-resolution synthetic input resampled to $1$~mm$^3$ using trilinear interpolation.

\subsection{Ablation experiments}
\label{sec:ablation}
% # Losses: mae vs wmae vs wmae + fft vs wmae + fft + dice @ 1 modelo
To quantify the contribution of each term in our composite loss function, we performed an ablation study using a single model for all studied resolutions on T1w data. Our baseline reconstruction relies on WMAE to not only penalize reconstruction errors but also recover sharp edges. We then incrementally integrated the spectral consistency term (FFT) and the semantic consistency term (Dice).

Tables \ref{tab:ablation_psnr} and \ref{tab:ablation_dice} summarize the quantitative impact of these components across resolution clusters using PSNR and Dice metrics. The addition of the spectral loss yielded an improvement in signal fidelity, raising the overall PSNR (Peak Signal-to-Noise Ratio) from $37.49$ dB to $37.93$ dB. These gains were more pronounced in lower resolution regimes where image degradation is most severe. This confirms that explicitly penalizing errors in the Fourier domain compels the network to recover high-frequency textural details that are often smoothed out by pixel-wise metrics alone. This spectral penalisation acts as a computationally efficient surrogate for perceptual loss, as each Fourier coefficient encodes global spatial correlations across the entire volume.

A higher performance gain was observed upon integrating the segmentation-based loss. The full composite loss ($\mathcal{L}_{WMAE}+\mathcal{L}_{FFT}+\mathcal{L}_{seg}$) achieved the highest performance across all metrics, reaching an overall PSNR of $38.15$ dB and a Dice score of $0.877$. While the numerical increase in Dice score (from $0.875$ to $0.877$) appears modest, it reflects critical structural corrections, as evidenced by the consistent gains in the most challenging low-resolution clusters ($6.5-10.5 \text{ mm}^3$) where anatomical ambiguity is highest.

\begin{table}[H]
\centering
\caption{\textbf{Quantitative ablation study of PSNR (mean $\pm$ standard deviation).} Impact of incremental loss terms on signal reconstruction ($\mathcal{L}_{WMAE}$ and $\mathcal{L}_{FFT} + \mathcal{L}_{seg}$). \textbf{Bold} indicates best performance. $^*$ indicates statistical significance ($p<0.05$) compared to the previous best result in the last overall column.}
\label{tab:ablation_psnr}
\resizebox{\columnwidth}{!}{%
\begin{tabular}{lcccccccc}
\toprule
\textbf{Architecture} & \textbf{1.0-1.5} & \textbf{1.5-2.5} & \textbf{2.5-3.5} & \textbf{3.5-4.5} & \textbf{4.5-5.5} & \textbf{5.5-6.5} & \textbf{6.5-10.5} & \textbf{Overall} \\
 & ($\mathrm{mm}^3$) & ($\mathrm{mm}^3$) & ($\mathrm{mm}^3$) & ($\mathrm{mm}^3$) & ($\mathrm{mm}^3$) & ($\mathrm{mm}^3$) & ($\mathrm{mm}^3$) & (Mean $\pm$ Std) \\
\midrule
Reference (Input) & $32.187 \pm 1.852$ & $31.448 \pm 1.865$ & $30.225 \pm 2.081$ & $29.872 \pm 1.923$ & $29.114 \pm 1.967$ & $28.897 \pm 1.988$ & $27.940 \pm 1.989$ & $29.600 \pm 2.300$ \\
\midrule
Baseline ($\mathcal{L}_{WMAE}$) & $39.843 \pm 1.781$ & $40.197 \pm 2.155$ & $38.416 \pm 2.181$ & $37.880 \pm 2.135$ & $36.912 \pm 2.082$ & $36.399 \pm 2.122$ & $35.399 \pm 1.979$ & $37.485 \pm 2.584$ \\
+ Frequency ($\mathcal{L}_{FFT}$) & $39.916 \pm 1.937$ & $40.030 \pm 1.902$ & $38.935 \pm 2.549$ & $38.432 \pm 2.190$ & $37.284 \pm 2.285$ & $37.084 \pm 2.409$ & $36.015 \pm 2.279$ & $37.927 \pm 2.627^*$ \\
\textbf{+ Semantic ($\mathcal{L}_{seg}$)} & $\mathbf{40.396 \pm 2.338}$ & $\mathbf{40.488 \pm 2.270}$ & $\mathbf{39.189 \pm 2.885}$ & $\mathbf{38.712 \pm 2.461}$ & $\mathbf{37.432 \pm 2.337}$ & $\mathbf{37.230 \pm 2.492}$ & $\mathbf{36.055 \pm 2.296}$ & $\mathbf{38.148 \pm 2.854}^*$ \\
\bottomrule
\end{tabular}
}
\end{table}

\begin{table}[H]
\centering
\caption{\textbf{Quantitative ablation study of Dice score (mean $\pm$ standard deviation).} Impact of incremental loss terms on tissue segmentation accuracy ($\mathcal{L}_{WMAE}$ and $\mathcal{L}_{FFT} + \mathcal{L}_{seg}$). \textbf{Bold} indicates best performance. $^*$ indicates statistical significance ($p<0.05$) compared to the previous best result in the last overall column.}
\label{tab:ablation_dice}
\resizebox{\columnwidth}{!}{%
\begin{tabular}{lcccccccc}
\toprule
\textbf{Architecture} & \textbf{1.0-1.5} & \textbf{1.5-2.5} & \textbf{2.5-3.5} & \textbf{3.5-4.5} & \textbf{4.5-5.5} & \textbf{5.5-6.5} & \textbf{6.5-10.5} & \textbf{Overall} \\
 & ($\mathrm{mm}^3$) & ($\mathrm{mm}^3$) & ($\mathrm{mm}^3$) & ($\mathrm{mm}^3$) & ($\mathrm{mm}^3$) & ($\mathrm{mm}^3$) & ($\mathrm{mm}^3$) & (Mean $\pm$ Std) \\
\midrule
Reference (Input) & $0.884 \pm 0.030$ & $0.883 \pm 0.029$ & $0.876 \pm 0.031$ & $0.873 \pm 0.029$ & $0.866 \pm 0.030$ & $0.863 \pm 0.032$ & $0.855 \pm 0.032$ & $0.869 \pm 0.032$ \\
\midrule
Baseline ($\mathcal{L}_{WMAE}$) & $0.886 \pm 0.031$ & $0.885 \pm 0.030$ & $0.880 \pm 0.032$ & $0.878 \pm 0.030$ & $0.871 \pm 0.032$ & $0.868 \pm 0.034$ & $0.860 \pm 0.034$ & $0.873 \pm 0.033$ \\
+ Frequency ($\mathcal{L}_{FFT}$) & $0.887 \pm 0.031$ & $0.887 \pm 0.030$ & $0.881 \pm 0.032$ & $0.879 \pm 0.030$ & $0.873 \pm 0.031$ & $0.870 \pm 0.033$ & $0.862 \pm 0.034$ & $0.875 \pm 0.033^*$ \\
\textbf{+ Semantic ($\mathcal{L}_{seg}$)} & $\mathbf{0.889 \pm 0.031}$ & $\mathbf{0.889 \pm 0.031}$ & $\mathbf{0.883 \pm 0.032}$ & $\mathbf{0.881 \pm 0.031}$ & $\mathbf{0.875 \pm 0.031}$ & $\mathbf{0.872 \pm 0.033}$ & $\mathbf{0.864 \pm 0.033}$ & $\mathbf{0.877 \pm 0.033}^*$ \\
\bottomrule
\end{tabular}
}
\end{table}

% # Ensembling? 

% # Best loses -> one model vs one model per cluster vs vol-aniso 
Once established the optimal composite loss function ($\mathcal{L}_{WMAE} + \mathcal{L}_{FFT} + \mathcal{L}_{seg}$), we evaluated the structural capacity of the network architectures to handle the diverse clinical degradation manifold. For that purpose, we compared the unified "Generalist" model against two MoE strategies: a clustering approach based solely on Volumetric Resolution (as done in early iterations of REMIX), and our proposed bivariate clustering approach that incorporates both Volumetric Resolution and Anisotropy. 

Table \ref{tab:moe_comparison} details the quantitative performance across the resolution spectrum. While the Generalist model achieves a competitive baseline (Overall: 38.148 dB), partitioning the problem space yields substantial gains. Moving to a volume-based MOE provides a moderate improvement (38.472 dB), primarily benefiting the higher-resolution clusters (1.0–2.5 mm$^3$). However, it struggles to significantly improve highly degraded regimes, indicating that volumetric resolution alone is an insufficient routing criterion.

\begin{figure}[H]
    \centering
    \includegraphics[width=\linewidth]{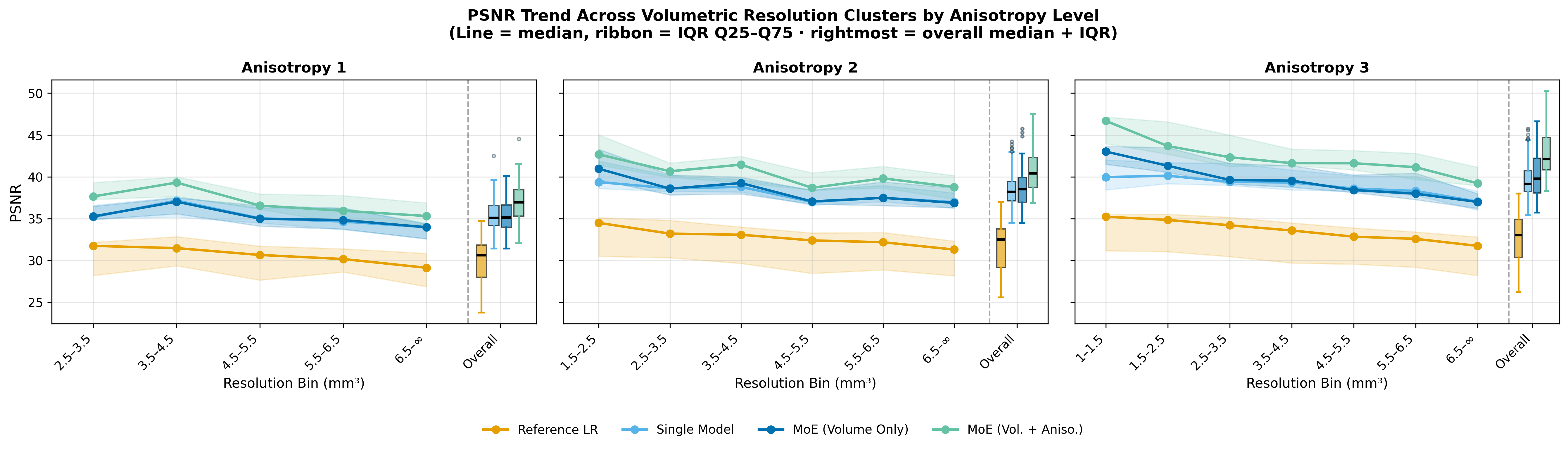}
    \caption{PSNR trend across the two-dimensional acquisition manifold. Anisotropy is mapped into three discrete clusters based on the voxel dimension ratio ($r_{\text{min}} / r_{\text{max}}$): Bin 1 (0 to 0.33), Bin 2 (0.33 to 0.66), and Bin 3 (0.66 to 1.0). Volumetric clusters are in shown in horizontal axis. }
    \label{fig:trendlines_psnr}
\end{figure}

The introduction of the proposed bivariate MoE (Volume + Anisotropy) produces a performance gain across all clusters, achieving an overall PSNR of 40.484 dB. Figure \ref{fig:trendlines_psnr} visualizes this performance trend across the entire two-dimensional acquisition manifold. 

As illustrated, the bivariate MoE (green line) maintain a strict, robust superiority over both the generalist ablation models (blue lines) across every volume and anisotropy cluster. By explicitly separating isotropic blur from highly anisotropic thick-slice acquisitions, the specialized expert networks can dedicate their full capacity to solve specific directional point spread functions (PSF). This architectural refinement yields an average improvement of over 2.3 dB compared to the generalist model, confirming that anisotropy is a critical confounding factor that unified networks struggle to resolve. Regarding computational complexity, while this approach increases the total burden during the training phase, scaling linearly with the number of experts (e.g., 5 hours for a single model, roughly 40 hours for 7 models, and 108 hours for 18 models) the framework incurs zero additional computational overhead during inference. Since the gating mechanism deterministically selects a single expert based on acquisition metadata, the run-time complexity remains identical to that of a standard single model ($O(1)$), ensuring rapid and efficient deployment in clinical workflows.

\begin{table}[H]
\centering
\caption{\textbf{Comparison of PSNR (mean $\pm$ standard deviation).} Bold indicates best performance. $^*$ indicates statistical significance ($p<0.001$) compared to the previous best result in the last overall column.}
\label{tab:moe_comparison}
\resizebox{\columnwidth}{!}{%
\begin{tabular}{lcccccccc}
\toprule
\textbf{Method} & \textbf{1.0-1.5} & \textbf{1.5-2.5} & \textbf{2.5-3.5} & \textbf{3.5-4.5} & \textbf{4.5-5.5} & \textbf{5.5-6.5} & \textbf{6.5-10.5} & \textbf{Overall} \\
 & ($\mathrm{mm}^3$) & ($\mathrm{mm}^3$) & ($\mathrm{mm}^3$) & ($\mathrm{mm}^3$) & ($\mathrm{mm}^3$) & ($\mathrm{mm}^3$) & ($\mathrm{mm}^3$) & (Mean $\pm$ Std) \\
\midrule
Reference (Input) & $32.187 \pm 1.852$ & $31.448 \pm 1.865$ & $30.225 \pm 2.081$ & $29.872 \pm 1.923$ & $29.114 \pm 1.967$ & $28.897 \pm 1.988$ & $27.940 \pm 1.989$ & $29.600 \pm 2.300$ \\
Generalist (Single Model) & $40.396 \pm 2.338$ & $40.488 \pm 2.270$ & $39.189 \pm 2.885$ & $38.712 \pm 2.461$ & $37.432 \pm 2.337$ & $37.230 \pm 2.492$ & $36.055 \pm 2.296$ & $38.148 \pm 2.854^*$ \\
MoE (Volume Only) & $42.997 \pm 2.030$ & $41.772 \pm 2.333$ & $39.056 \pm 2.749$ & $38.875 \pm 2.347$ & $37.354 \pm 2.362$ & $37.243 \pm 2.535$ & $36.279 \pm 2.434$ & $38.472 \pm 3.114$ \\
\textbf{MoE (Vol. + Aniso.)} & $\mathbf{45.748 \pm 2.416}$ & $\mathbf{43.870 \pm 2.570}$ & $\mathbf{40.947 \pm 3.106}$ & $\mathbf{41.024 \pm 2.546}$ & $\mathbf{39.204 \pm 3.213}$ & $\mathbf{39.148 \pm 3.182}$ & $\mathbf{38.239 \pm 2.894}$ & $\mathbf{40.484 \pm 3.585}^*$ \\
\bottomrule
\end{tabular}
}
\end{table}

\subsection{Comparison with state-of-the-art methods}
\label{sec:comparison}

The problem formulation described in Section \ref{subsec:formulation} has an important consequence for the choice of comparable methods. Self-supervised approaches such as SMORE~\parencite{Remedios2023Self-SupervisedGap} exploit high-resolution in-plane information to super-resolve exclusively along the thick-slice axis, leaving the in-plane resolution unchanged. This design is well-suited to anisotropic acquisitions whose in-plane pixel size already matches the desired isotropic target, but it fails to generalise to the heterogeneous clinical scenario addressed here. Furthermore, SMORE assumes a strictly anisotropic input with a clearly defined high-quality reference plane; for the near-isotropic degradations present in our evaluation manifold ($A_3$ bin, $A \in (0.66,\,1.0]$), no such reference plane exists and the method provides no mechanism to act. Arbitrary-scale methods such as ArSSR~\parencite{Wu2023} learn a continuous implicit neural representation that maps any input coordinate to its high-resolution counterpart at a user-specified scale factor. By construction, this scale factor is applied independently and uniformly to each spatial axis; the method does not model the physical constraint that, for a pre-resampled $1$~mm$^3$ input, the in-plane axes require no further scaling while only the through-plane axis needs residual high-frequency recovery.

Accordingly, comparisons are restricted to methods that share this output specification, namely SynthSR \parencite{Iglesias2021JointContrast} and SuperSynth \parencite{Liu2025AImaging}. REMIX~\parencite{Morell-Ortega2024REMIX:Experts}, our prior framework, is additionally included as an intermediate reference. Since 
REMIX operates in MNI152 standard space, evaluation inputs were resampled to the MNI152 $1$~mm$^3$ isotropic grid (181 x 217 x 181 voxels) prior to inference, and the enhanced outputs were subsequently resampled back to native space for metric computation against the native-space ground truth. The implications of this resampling round-trip are discussed in the context of each metric below.

Quantitative results are reported across four complementary metrics: PSNR, correlation coefficient (CC), Learned Perceptual Image Patch Similarity (LPIPS) ~\parencite{Zhang2018} and Dice score.

Table \ref{tab:sota_comparison} reveals a consistent pattern across all seven 
degradation bins: both SynthSR ($25.074 \pm 2.042$~dB) and SuperSynth 
($28.490 \pm 2.271$~dB) fall below the unprocessed low-resolution reference 
($29.600 \pm 2.300$~dB), indicating that label-based generative enhancement 
actively degrades signal fidelity relative to the input. Generative methods optimise for distributional plausibility rather than voxel-level reconstruction, and the synthesis process introduces a predefined contrast profile derived from the training label distribution that need not align with the true intensity range of the input acquisition. To ensure a fair comparison, the mean intensity of all generative outputs was 
normalised to the ground truth prior to evaluation; the gap persists regardless, 
confirming that it reflects structural divergence rather than a scalar 
offset artifact. REMIX ($30.127 \pm 2.512$~dB) substantially outperforms 
both generative baselines, confirming that residual 
reconstruction is inherently more signal-faithful than label-based synthesis 
regardless of generation quality. CAHAL ($40.484 \pm 3.585$~dB) achieves a 
further gain of $10.3$~dB over REMIX, the largest single 
improvement in the table, driven by the combination of native-space processing 
and bivariate expert specialisation.

\begin{table}[H]
\centering
\caption{\textbf{Comparison of PSNR (mean $\pm$ standard deviation).} Bold indicates best performance. $^*$ indicates statistical significance ($p<0.001$) compared to the previous best result in the last overall column.}
\label{tab:sota_comparison}
\resizebox{\columnwidth}{!}{%
\begin{tabular}{lcccccccc}
\toprule
\textbf{Method} & \textbf{1.0-1.5} & \textbf{1.5-2.5} & \textbf{2.5-3.5} & \textbf{3.5-4.5} & \textbf{4.5-5.5} & \textbf{5.5-6.5} & \textbf{6.5-10.5} & \textbf{Overall} \\
 & ($\mathrm{mm}^3$) & ($\mathrm{mm}^3$) & ($\mathrm{mm}^3$) & ($\mathrm{mm}^3$) & ($\mathrm{mm}^3$) & ($\mathrm{mm}^3$) & ($\mathrm{mm}^3$) & (Mean $\pm$ Std) \\
\midrule
Reference (Input) & $32.187 \pm 1.852$ & $31.448 \pm 1.865$ & $30.225 \pm 2.081$ & $29.872 \pm 1.923$ & $29.114 \pm 1.967$ & $28.897 \pm 1.988$ & $27.940 \pm 1.989$ & $29.600 \pm 2.300$ \\
\midrule
\multicolumn{9}{l}{\textit{State-of-the-Art (Generative)}} \\
SynthSR \parencite{Iglesias2021JointContrast} & $25.105 \pm 2.096$ & $25.187 \pm 2.051$ & $25.220 \pm 2.009$ & $25.095 \pm 2.053$ & $25.120 \pm 2.019$ & $24.943 \pm 2.058$ & $24.927 \pm 2.024$ & $25.074 \pm 2.042^*$ \\
SuperSynth \parencite{Liu2025AImaging} & $28.702 \pm 2.520$ & $28.686 \pm 2.374$ & $28.644 \pm 2.414$ & $28.409 \pm 2.297$ & $28.488 \pm 2.115$ & $28.304 \pm 2.284$ & $28.432 \pm 2.072$ & $28.490 \pm 2.271^*$ \\
\midrule
\multicolumn{9}{l}{\textit{Prior Work (Residual, MNI space)}} \\ \\
REMIX \parencite{Morell-Ortega2024REMIX:Experts}
    & $35.180 \pm 1.199$ & $34.593 \pm 1.221$ & $30.126 \pm 1.417$ & $29.532 \pm 1.422$
    & $28.724 \pm 1.452$ & $28.468 \pm 0.752$ & $29.175 \pm 1.266$ & $30.127 \pm 2.512^*$ \\
\midrule
\multicolumn{9}{l}{\textit{Proposed Framework (Residual, native space)}} \\
\textbf{CAHAL} & $\mathbf{45.748 \pm 2.416}$ & $\mathbf{43.870 \pm 2.570}$ & $\mathbf{40.947 \pm 3.106}$ & $\mathbf{41.024 \pm 2.546}$ & $\mathbf{39.204 \pm 3.213}$ & $\mathbf{39.148 \pm 3.182}$ & $\mathbf{38.239 \pm 2.894}$ & $\mathbf{40.484 \pm 3.585}^*$ \\
\bottomrule
\end{tabular}
}
\end{table}

However, PSNR is sensitive to the absolute intensity range, and a residual dynamic-range advantage for REMIX and CAHAL, both of which operate entirely within the native image intensity domain, must be accounted for. We therefore complement PSNR with 
two metrics that are by construction insensitive to this confound: the Pearson 
Correlation Coefficient (CC), which evaluates structural fidelity independently 
of global intensity scale, and LPIPS, a perceptual 
distance measure that operates in a learned feature space and is entirely 
unaffected by intensity range differences as the inputs are normalized equally.

The CC results, reported in Table \ref{tab:architecture_comparison_cc_stats}, 
 corroborate the PSNR findings. SynthSR ($0.979 \pm 0.004$) falls below the low-resolution reference ($0.983 \pm 0.014$), confirming that its synthesis process introduces systematic structural distortions sufficiently large to disrupt global 
intensity correlations. SuperSynth ($0.990 \pm 0.003$) recovers to match,  
and marginally exceed, the reference, indicating that its more recent 
generative architecture succeeds in preserving broad structural relationships 
even if voxel-level fidelity remains compromised. REMIX ($0.990 \pm 0.004$) 
achieves an identical overall CC to SuperSynth, and the difference between 
them is not statistically significant. This 
convergence is chronologically coherent: although the two methods follow 
entirely different paradigms, residual reconstruction versus generative 
synthesis, both represent successive generations of the state of the art 
published within a year of each other, and their structural correlation 
performance has reached the same ceiling at this metric level. CAHAL ($0.998 \pm 0.002$) breaks through this ceiling 
with a statistically significant improvement over all preceding methods, approaching a near-perfect structural agreement with the 
ground truth and maintaining near-zero variance across all seven bins, 
confirming consistent enhancement quality across the full degradation manifold 
under a metric entirely free of intensity range confounds.

\begin{table}[H]
\centering
\caption{\textbf{Comparison of CC (mean $\pm$ standard deviation).} \textbf{Bold} indicates best performance. $^*$ indicates statistical significance ($p<0.001$) compared to the previous best result in the last Overall column; $^{\mathrm{ns}}$ indicates 
no significant difference from the previous best.}
\label{tab:architecture_comparison_cc_stats}
\resizebox{\columnwidth}{!}{%
\begin{tabular}{lcccccccc}
\toprule
\textbf{Method} & \textbf{1.0-1.5} & \textbf{1.5-2.5} & \textbf{2.5-3.5} & \textbf{3.5-4.5} & \textbf{4.5-5.5} & \textbf{5.5-6.5} & \textbf{6.5-10.5} & \textbf{Overall} \\
 & ($\mathrm{mm}^3$) & ($\mathrm{mm}^3$) & ($\mathrm{mm}^3$) & ($\mathrm{mm}^3$) & ($\mathrm{mm}^3$) & ($\mathrm{mm}^3$) & ($\mathrm{mm}^3$) & (Mean $\pm$ Std) \\
\midrule
Reference (Input)  & $0.990 \pm 0.008$ & $0.989 \pm 0.009$ & $0.986 \pm 0.011$ & $0.984 \pm 0.012$ & $0.982 \pm 0.015$ & $0.980 \pm 0.015$ & $0.977 \pm 0.018$ & $0.983 \pm 0.014$ \\
\midrule
\multicolumn{9}{l}{\textit{State-of-the-Art (Generative)}} \\

SynthSR \parencite{Iglesias2021JointContrast}& $0.980 \pm 0.004$ & $0.980 \pm 0.004$ & $0.980 \pm 0.004$ & $0.980 \pm 0.004$ & $0.979 \pm 0.004$ & $0.979 \pm 0.004$ & $0.978 \pm 0.004$ & $0.979 \pm 0.004 ^*$ \\
SuperSynth \parencite{Liu2025AImaging} & $0.991 \pm 0.003$ & $0.991 \pm 0.003$ & $0.990 \pm 0.003$ & $0.990 \pm 0.003$ & $0.990 \pm 0.003$ & $0.990 \pm 0.003$ & $0.989 \pm 0.003$ & $0.990 \pm 0.003^*$ \\
\midrule
\multicolumn{9}{l}{\textit{Prior Work (Residual, MNI space)}} \\ \\
REMIX \parencite{Morell-Ortega2024REMIX:Experts}
    & $0.996 \pm 0.002$ & $0.996 \pm 0.002$ & $0.993 \pm 0.002$ & $0.990 \pm 0.002$
    & $0.989 \pm 0.002$ & $0.986 \pm 0.001$ & $0.990 \pm 0.002$ & $0.990 \pm 0.004^{\mathrm{ns}}$ \\
\midrule
\multicolumn{9}{l}{\textit{Proposed Framework (Residual, native space)}} \\
\textbf{CAHAL} & $\mathbf{0.999 \pm 0.000}$ & $\mathbf{0.999 \pm 0.000}$ & $\mathbf{0.999 \pm 0.001}$ & $\mathbf{0.998 \pm 0.001}$ & $\mathbf{0.998 \pm 0.001}$ & $\mathbf{0.997 \pm 0.002}$ & $\mathbf{0.997 \pm 0.002}$ & $\mathbf{0.998 \pm 0.002^*}$ \\
\hline
\end{tabular}
}
\end{table}

To confirm that the advantage of residual methods over generative baselines is not an artefact of intensity range differences, Table~\ref{tab:lpips} reports LPIPS scores \parencite{Zhang2018}. Lower values indicate greater 
perceptual similarity with the ground truth.

SynthSR ($0.266 \pm 0.067$) is substantially worse than the low-resolution reference ($0.181 \pm 0.068$), confirming that its outputs are perceptually dissimilar from the true anatomy even under a 
scale-invariant criterion. SuperSynth ($0.174 \pm 0.065$) is the only competing method to match the reference perceptually. However, this result exposes a fundamental dissociation: SuperSynth achieves perceptual plausibility while falling below the reference on every structural metric, confirming that an image can look correct without preserving the underlying anatomy, and that perceptual metrics alone are insufficient proxies for clinical safety. REMIX ($0.208 \pm 0.078$) shows a monotonically worsening LPIPS from Bin~1 ($0.116$) to Bin~7 ($0.252$), independently corroborating the MNI-space interpolation degradation. CAHAL ($0.058 \pm 0.048$) outperforms all methods across every bin, with the margin widening at lower degradation bins where the residual correction required is smallest and most precisely learnable.

\begin{table}[H]
\centering
\caption{\textbf{Comparison of LPIPS (mean $\pm$ standard deviation), 
RadNet backbone.} Lower values indicate greater perceptual similarity 
to the ground truth. \textbf{Bold} indicates best (lowest) performance. 
$^*$ indicates statistical significance ($p<0.001$) compared to the 
previous best result in the Overall column; $^{\mathrm{ns}}$ indicates 
no significant difference from the previous best.}
\label{tab:lpips}
\resizebox{\columnwidth}{!}{%
\begin{tabular}{lcccccccc}
\toprule
\textbf{Method} & \textbf{1.0--1.5} & \textbf{1.5--2.5} & \textbf{2.5--3.5} 
    & \textbf{3.5--4.5} & \textbf{4.5--5.5} & \textbf{5.5--6.5} 
    & \textbf{6.5--10.5} & \textbf{Overall} \\
 & ($\mathrm{mm}^3$) & ($\mathrm{mm}^3$) & ($\mathrm{mm}^3$) 
 & ($\mathrm{mm}^3$) & ($\mathrm{mm}^3$) & ($\mathrm{mm}^3$) 
 & ($\mathrm{mm}^3$) & (Mean $\pm$ Std) \\
\midrule
Reference (Input) 
    & $0.105 \pm 0.032$ & $0.125 \pm 0.037$ & $0.159 \pm 0.049$ 
    & $0.169 \pm 0.053$ & $0.196 \pm 0.056$ & $0.206 \pm 0.074$ 
    & $0.234 \pm 0.067$ & $0.181 \pm 0.068$ \\
\midrule
\multicolumn{9}{l}{\textit{State-of-the-Art (Generative)}} \\
SynthSR \parencite{Iglesias2021JointContrast}
    & $0.268 \pm 0.066$ & $0.263 \pm 0.067$ & $0.264 \pm 0.065$ 
    & $0.264 \pm 0.067$ & $0.267 \pm 0.068$ & $0.267 \pm 0.068$ 
    & $0.268 \pm 0.067$ & $0.266 \pm 0.067^*$ \\
SuperSynth \parencite{Liu2025AImaging}
    & $0.156 \pm 0.060$ & $0.156 \pm 0.081$ & $0.165 \pm 0.060$ 
    & $0.169 \pm 0.063$ & $0.179 \pm 0.063$ & $0.182 \pm 0.070$ 
    & $0.190 \pm 0.066$ & $0.174 \pm 0.065^{\mathrm{ns}}$ \\
\midrule
\multicolumn{9}{l}{\textit{Prior Work (Residual, MNI space)}} \\ \\
REMIX \parencite{Morell-Ortega2024REMIX:Experts}
    & $0.116 \pm 0.042$ & $0.132 \pm 0.045$ & $0.175 \pm 0.053$ 
    & $0.207 \pm 0.066$ & $0.233 \pm 0.066$ & $0.249 \pm 0.081$ 
    & $0.252 \pm 0.067$ & $0.208 \pm 0.078^*$ \\
\midrule
\multicolumn{9}{l}{\textit{Proposed Framework (Residual, native space)}} \\
\textbf{CAHAL}
    & $\mathbf{0.026 \pm 0.024}$ & $\mathbf{0.031 \pm 0.028}$ & $\mathbf{0.048 \pm 0.035}$ 
    & $\mathbf{0.051 \pm 0.042}$ & $\mathbf{0.064 \pm 0.043}$ & $\mathbf{0.075 \pm 0.057}$ 
    & $\mathbf{0.082 \pm 0.055}$ & $\mathbf{0.058 \pm 0.048}^*$ \\
\bottomrule
\end{tabular}
}
\end{table}

% Across all three signal fidelity metrics, the results reveal a consistent 
% structural divide between the two methodological paradigms evaluated here. 
% Generative methods optimise for distributional plausibility, synthesising 
% high-frequency detail that matches the appearance of the training label 
% distribution rather than the patient's true underlying anatomy. This approach 
% produces visually compelling outputs but introduces systematic deviations from 
% the native signal, deviations that are detectable as reduced PSNR, degraded 
% CC, and, as shown below, impaired segmentation accuracy. Residual methods, 
% by contrast, constrain the network to learn only the high-frequency correction 
% required to restore the input: the global residual connection ensures that 
% the output remains anchored to the original native signal, making the network 
% structurally incapable of replacing anatomical structures with synthesised 
% alternatives. This distinction has direct clinical implications in the presence 
% of pathology, lesions, atrophy patterns, or structural anomalies absent from 
% the training label distribution will be preserved by a residual network and 
% may be inpainted by a generative one. Furthermore, by explicitly modelling 
% the degradation physics through the bivariate expert partitioning, CAHAL 
% recovers the underlying signal with a specificity that a single generalist 
% model, whether generative or residual, cannot match across the full range 
% of clinical acquisition protocols

\begin{figure}[H]
    \centering
    \includegraphics[width=\textwidth]{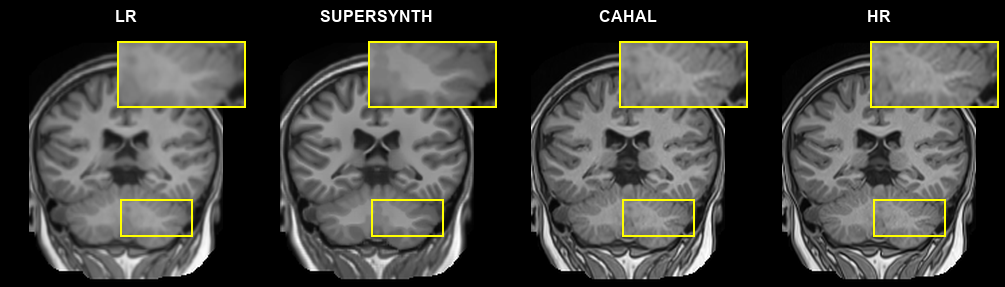}
    \caption{Qualitative assessment of cerebellar reconstruction of a case from volumetric cluster 5 anisotropy 1 (1x1x5 $\text{mm}^3$). CAHAL better reconstructs the LR image than SuperSynth, especially in the cerebellum.}
    \label{fig:qualitative_results_t1}
\end{figure}

This philosophical difference is explicitly corroborated by our qualitative results (Figure \ref{fig:qualitative_results_t1} and Appendix Figure \ref{fig:deepceres_example}). Visually, while SuperSynth effectively removes blur and generates high-contrast, visually pleasing MRI volumes, closer inspection reveals that it frequently alters subtle cortical folding patterns, removes textures and hallucinates tissue boundaries that do not exist in the High-Resolution (HR) ground truth. Conversely, CAHAL successfully sharpens the image while strictly maintaining the topological features of the patient's true anatomy. The clinical danger of these generative hallucinations will be further validated in the following sections. 

% imagen cualitativa de los resultados para distintos R

% While signal fidelity is indicative of reconstruction quality, clinical utility is best measured by downstream task performance. Table \ref{tab:sota_comparison_dice} presents the mean segmentation accuracy (Dice score) on 8 distinct tissue classes. It is crucial to note that these tissue semantic definitions align with the supervision signals used during the training of both the generative baselines (SynthSR, SuperSynth) and our proposed framework. Here, the risks of unconstrained generation become apparent. SynthSR ($0.813$) and SuperSynth ($0.855$) result in lower segmentation accuracy compared to the original Low-Resolution input ($0.869$), suggesting that while the images look sharper, the structural boundaries may have been distorted. Conversely, our bivariate MoE framework ($0.886$) not only preserves the original structure but significantly improves segmentation accuracy over the input ($p < 0.001$). This advantage is most pronounced in the near-isotropic regime ($1.0-1.5 \text{ mm}^3$), where our method achieves a Dice score of $0.901$ compared to SuperSynth's $0.860$. This confirms that for clinical super-resolution, where diagnostic accuracy depends on the faithful preservation of existing structures, our specialized residual formulation offers a safer and more reliable solution than purely generative synthesis.

While signal fidelity metrics assess reconstruction quality in isolation, clinical 
utility ultimately depends on whether enhanced images support reliable downstream 
analysis. Table \ref{tab:sota_comparison_dice} reports mean Dice score across eight 
tissue classes, evaluating segmentation accuracy on the enhanced outputs relative 
to native-space ground truth labels. 

Both SynthSR ($0.813 \pm 0.024$) and SuperSynth ($0.855 \pm 0.030$) fall below the unprocessed low-resolution reference ($0.869 \pm 0.032$), confirming that visually sharper images do not translate to more anatomically faithful ones. This outcome is consistent with the hallucination risk identified by \textcite{Cohen2018DistributionTranslation}.

REMIX~\parencite{Morell-Ortega2024REMIX:Experts} presents a qualitatively 
different failure profile that is equally instructive. Despite adopting the same 
residual reconstruction paradigm as CAHAL, REMIX achieves an overall Dice of 
$0.834 \pm 0.028$, below both the low-resolution reference and SuperSynth,
with the deficit growing markedly at higher anisotropy bins (Bin~6: $0.799$, compared 
to the reference $0.863$). This degradation is consequence of the spatial framework in which REMIX operates. For near-isotropic acquisitions (Bin~1--2), the forward and backward interpolation introduces 
only modest geometric distortion, and REMIX ($0.866$) remains within a narrow margin of the reference ($0.884$). For highly anisotropic inputs the spatial errors propagate through. The gradient of this effect 
across bins is visible directly in Table~\ref{tab:sota_comparison_dice} and constitutes 
the primary empirical motivation for CAHAL's native-space design. Performing 
enhancement before any registration step eliminates this error cascade entirely.

CAHAL ($0.887 \pm 0.035$) is the only method to consistently exceed the low-resolution 
reference across all seven bins, achieving statistically significant improvement over 
the reference and over all competing methods. The advantage is most 
pronounced in the near-isotropic regime (Bin~1: $0.902$ vs.\ SuperSynth $0.860$), 
where the expert network operates closest to its training optimum, but remains 
consistent through Bin~7 ($0.875$ vs.\ reference $0.855$), demonstrating that 
the bivariate specialisation maintains reconstruction safety even under severe 
anisotropic degradation.

\begin{table}[H]
\centering
\caption{\textbf{Comparison of Dice Score (mean $\pm$ standard deviation).} \textbf{Bold} indicates best performance. $^*$ indicates statistical significance ($p<0.001$) compared to the previous best result in the last overall column.}
\label{tab:sota_comparison_dice}
\resizebox{\columnwidth}{!}{%
\begin{tabular}{lcccccccc}
\toprule
\textbf{Method} & \textbf{1.0-1.5} & \textbf{1.5-2.5} & \textbf{2.5-3.5} & \textbf{3.5-4.5} & \textbf{4.5-5.5} & \textbf{5.5-6.5} & \textbf{6.5-10.5} & \textbf{Overall} \\
 & ($\mathrm{mm}^3$) & ($\mathrm{mm}^3$) & ($\mathrm{mm}^3$) & ($\mathrm{mm}^3$) & ($\mathrm{mm}^3$) & ($\mathrm{mm}^3$) & ($\mathrm{mm}^3$) & (Mean $\pm$ Std) \\
\midrule
Reference (Input) & $0.884 \pm 0.030$ & $0.883 \pm 0.029$ & $0.876 \pm 0.031$ & $0.873 \pm 0.029$ & $0.866 \pm 0.030$ & $0.863 \pm 0.032$ & $0.855 \pm 0.032$ & $0.869 \pm 0.032$ \\
\midrule
\multicolumn{9}{l}{\textit{State-of-the-Art (Generative)}} \\
SynthSR \parencite{Iglesias2021JointContrast} & $0.819 \pm 0.025$ & $0.818 \pm 0.024$ & $0.814 \pm 0.024$ & $0.816 \pm 0.023$ & $0.813 \pm 0.023$ & $0.811 \pm 0.024$ & $0.809 \pm 0.023$ & $0.813 \pm 0.024$ \\
SuperSynth \parencite{Liu2025AImaging} & $0.860 \pm 0.032$ & $0.860 \pm 0.031$ & $0.857 \pm 0.031$ & $0.856 \pm 0.030$ & $0.854 \pm 0.029$ & $0.852 \pm 0.029$ & $0.849 \pm 0.027$ & $0.855 \pm 0.030$ \\
\midrule
\multicolumn{9}{l}{\textit{Prior Work (Residual, MNI space)}} \\ \\
REMIX \parencite{Morell-Ortega2024REMIX:Experts}
    & $0.866 \pm 0.007$ & $0.866 \pm 0.008$ & $0.851 \pm 0.014$ 
    & $0.837 \pm 0.014$ & $0.823 \pm 0.015$ & $0.799 \pm 0.027$ 
    & $0.831 \pm 0.025$ & $0.834 \pm 0.028$ \\
\midrule
\multicolumn{9}{l}{\textit{Proposed Framework (Residual, native space)}} \\
\textbf{CAHAL} & $\mathbf{0.901 \pm 0.037}$ & $\mathbf{0.897 \pm 0.036}$ & $\mathbf{0.889 \pm 0.033}$ & $\mathbf{0.890 \pm 0.033}$ & $\mathbf{0.882 \pm 0.035}$ & $\mathbf{0.881 \pm 0.036}$ & $\mathbf{0.875 \pm 0.034}$ & $\mathbf{0.886 \pm 0.035}^*$ \\
\bottomrule
\end{tabular}
}
\end{table}

% Resultados en Flair
To assess the generalisability of CAHAL beyond standard T1-weighted imaging, 
we trained a dedicated instance of the framework on the FLAIR modality. FLAIR acquisitions are critical for the assessment of white matter hyperintensities (e.g., in Multiple Sclerosis) but frequently suffer from low through-plane resolution in clinical protocols. SynthSR and REMIX are excluded from the FLAIR comparison as neither framework supports FLAIR synthesis.

We trained our two-dimensional MoE using the curated FLAIR dataset and compared performance against SuperSynth which also generates FLAIR images.

Results of the specific FLAIR trained models are summarized in Tables \ref{tab:flair_psnr} and \ref{tab:flair_dice}. As can be noted, the trends observed in T1-weighted imaging are strongly reinforced here. CAHAL achieves an overall PSNR of $38.22$ dB, significantly outperforming the reference input ($32.80$ dB). In contrast, SuperSynth ($29.19$ dB) fails to improve upon the low-resolution input in terms of signal fidelity. This suggests that while SuperSynth may enforce a stylistic match to the FLAIR domain, it does not strictly respect the voxel-wise anatomical constraints required for high-fidelity restoration. The divergence in clinical utility is even more pronounced. The generative nature of SuperSynth results in a degradation of segmentation performance ($0.864$) compared to the baseline LR input ($0.883$). This indicates that the generative process may be eroding fine tissue boundaries or lesion borders. Conversely, CAHAL achieves a substantial improvement, reaching an overall Dice score of $0.944$. This near-perfect preservation of tissue semantics ($+0.06$ over input) confirms that our method effectively restores the sharp transitions between lesions, cerebrospinal fluid, and normal appearing white matter, which are essential for accurate automated volumetry.

\begin{table}[h!]
\centering
\caption{\textbf{FLAIR Modality - PSNR (mean $\pm$ standard deviation).} Comparison of restoration quality on clinical FLAIR images. \textbf{Bold} indicates best performance. $^*$ indicates statistical significance ($p<0.001$) compared to the input in the last overall column.}
\label{tab:flair_psnr}
\resizebox{\columnwidth}{!}{%
\begin{tabular}{lcccccccc}
\toprule
\textbf{Method} & \textbf{1.0-1.5} & \textbf{1.5-2.5} & \textbf{2.5-3.5} & \textbf{3.5-4.5} & \textbf{4.5-5.5} & \textbf{5.5-6.5} & \textbf{6.5-10.5} & \textbf{Overall} \\
 & ($\mathrm{mm}^3$) & ($\mathrm{mm}^3$) & ($\mathrm{mm}^3$) & ($\mathrm{mm}^3$) & ($\mathrm{mm}^3$) & ($\mathrm{mm}^3$) & ($\mathrm{mm}^3$) & (Mean $\pm$ Std) \\
\midrule
Reference (Input) & $37.448 \pm 2.951$ & $34.624 \pm 1.677$ & $33.399 \pm 0.554$ & $32.473 \pm 0.577$ & $31.877 \pm 0.738$ & $31.114 \pm 0.688$ & $30.209 \pm 0.981$ & $32.799 \pm 2.445$ \\
\midrule
SuperSynth \parencite{Liu2025AImaging} & $30.013 \pm 0.424$ & $29.727 \pm 0.192$ & $29.444 \pm 0.282$ & $29.225 \pm 0.386$ & $29.080 \pm 0.468$ & $28.790 \pm 0.546$ & $28.327 \pm 0.925$ & $29.190 \pm 0.723$ \\
\textbf{CAHAL} & $\mathbf{43.391 \pm 2.092}$ & $\mathbf{40.833 \pm 1.256}$ & $\mathbf{39.432 \pm 0.810}$ & $\mathbf{37.891 \pm 1.407}$ & $\mathbf{36.916 \pm 1.253}$ & $\mathbf{36.138 \pm 1.493}$ & $\mathbf{34.686 \pm 2.137}$ & $\mathbf{38.224 \pm 3.006}^*$ \\
\bottomrule
\end{tabular}
}
\end{table}

\begin{table}[h!]
\centering
\caption{\textbf{FLAIR Modality - Segmentation Accuracy Dice (mean $\pm$ standard deviation).} Impact of super-resolution on multi-tissue segmentation in FLAIR images. \textbf{Bold} indicates best performance. $^*$ indicates statistical significance ($p<0.001$) compared to the input in the last overall column.}
\label{tab:flair_dice}
\resizebox{\columnwidth}{!}{%
\begin{tabular}{lcccccccc}
\toprule
\textbf{Method} & \textbf{1.0-1.5} & \textbf{1.5-2.5} & \textbf{2.5-3.5} & \textbf{3.5-4.5} & \textbf{4.5-5.5} & \textbf{5.5-6.5} & \textbf{6.5-10.5} & \textbf{Overall} \\
 & ($\mathrm{mm}^3$) & ($\mathrm{mm}^3$) & ($\mathrm{mm}^3$) & ($\mathrm{mm}^3$) & ($\mathrm{mm}^3$) & ($\mathrm{mm}^3$) & ($\mathrm{mm}^3$) & (Mean $\pm$ Std) \\
\midrule
Reference (Input) & $0.939 \pm 0.020$ & $0.914 \pm 0.015$ & $0.899 \pm 0.006$ & $0.883 \pm 0.014$ & $0.872 \pm 0.019$ & $0.856 \pm 0.023$ & $0.834 \pm 0.032$ & $0.883 \pm 0.037$ \\
\midrule
SuperSynth \parencite{Liu2025AImaging} & $0.880 \pm 0.005$ & $0.875 \pm 0.004$ & $0.871 \pm 0.005$ & $0.865 \pm 0.009$ & $0.861 \pm 0.011$ & $0.854 \pm 0.016$ & $0.844 \pm 0.021$ & $0.864 \pm 0.016$ \\
\textbf{CAHAL} & $\mathbf{0.975 \pm 0.005}$ & $\mathbf{0.966 \pm 0.006}$ & $\mathbf{0.958 \pm 0.006}$ & $\mathbf{0.946 \pm 0.013}$ & $\mathbf{0.938 \pm 0.018}$ & $\mathbf{0.928 \pm 0.023}$ & $\mathbf{0.909 \pm 0.034}$ & $\mathbf{0.944 \pm 0.028}^*$ \\
\bottomrule
\end{tabular}
}
\end{table}
% \begin{tabular}{lcccccccc}
% \hline
% Architecture & 1.0-1.5  $\mathrm{mm}^3$ & 1.5-2.5  $\mathrm{mm}^3$ & 2.5-3.5  $\mathrm{mm}^3$ & 3.5-4.5  $\mathrm{mm}^3$ & 4.5-5.5  $\mathrm{mm}^3$ & 5.5-6.5  $\mathrm{mm}^3$ & 6.5-10.5  $\mathrm{mm}^3$ & Overall \\
% \hline
% REF\_LR\_DENOISED & $32.187 \pm 1.852$ & $31.448 \pm 1.865$ & $30.225 \pm 2.081$ & $29.872 \pm 1.923$ & $29.114 \pm 1.967$ & $28.897 \pm 1.988$ & $27.940 \pm 1.989$ & $29.600 \pm 2.300$ \\

% ONE\_MODEL & $40.396 \pm 2.338$ & $40.488 \pm 2.270$ & $39.189 \pm 2.885$ & $38.712 \pm 2.461$ & $37.432 \pm 2.337$ & $37.230 \pm 2.492$ & $36.055 \pm 2.296$ & $38.148 \pm 2.854$ \\
% VOL\_RES\_ONLY & $42.997 \pm 2.030$ & $41.772 \pm 2.333$ & $39.056 \pm 2.749$ & $38.875 \pm 2.347$ & $37.354 \pm 2.362$ & $37.243 \pm 2.535$ & $36.279 \pm 2.434$ & $38.472 \pm 3.114$ \\
% UNET\_64F\_REC\_DICE & $45.748 \pm 2.416$ & $43.870 \pm 2.570$ & $40.947 \pm 3.106$ & $41.024 \pm 2.546$ & $39.204 \pm 3.213$ & $39.148 \pm 3.182$ & $38.239 \pm 2.894$ & $40.484 \pm 3.585$ \\
% \hline
% \end{tabular}

\begin{figure}[H]
    \centering
    \includegraphics[width=\textwidth]{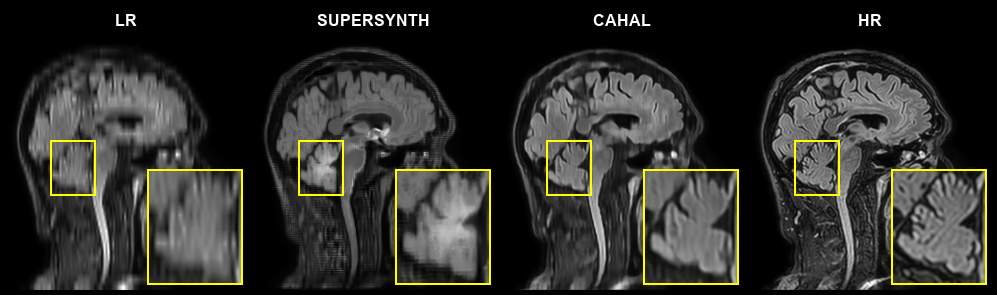}
    \caption{Example reconstruction case of 1x1x7 $\text{mm}^3$ with SuperSynth and CAHAL.}
    \label{fig:qualitative_results_flair}
\end{figure}

To assess practical deployability, we evaluated CAHAL's computational efficiency 
across a range of hardware configurations. On a 32~GB VRAM GPU, CAHAL processes 
a standard $200 \times 200 \times 200$ volume in approximately $3.5$ seconds, 
compared to $30$ seconds for SuperSynth under identical conditions. To support 
deployment on standard hardware, CAHAL incorporates a memory-aware tiled 
inference strategy: prior to processing, the framework queries available GPU VRAM 
and automatically selects the largest viable patch size, scaling the degree of 
tiling inversely with available memory. Total inference time remains under 
$10$ seconds across all GPU configurations from 8 to 32~GB, with reconstruction 
fidelity stable throughout confirming that boundary tiling introduces no measurable anatomical error. CPU execution is supported for environments without GPU access, though inference times increase considerably. 

The complete pipeline and pre-trained models are available to the scientific community at \href{https://github.com/volBrain-net/CAHAL}{https://github.com/volBrain-net/CAHAL}.

\subsection{Clinical validation and downstream task analysis}
\label{sec:validation}

Beyond the quantitative reconstruction metrics presented in the previous section, the ultimate utility of a super-resolution tool lies in its ability to facilitate robust clinical tasks, specifically, automated segmentation and volumetric analysis. Morphometric measurements, such as hippocampal volume, are fundamental for the early diagnosis and tracking of neurodegenerative pathologies like Alzheimer’s Disease (AD) \parencite{Frisoni2010TheDisease}. A parallel necessity exists in Multiple Sclerosis (MS), where accurate lesion volumetry serves as a critical surrogate marker for monitoring disease burden, progression, and therapeutic efficacy \parencite{Wattjes20212021Sclerosis, Filippi2016MRIGuidelines}.

Consequently, any enhancement method must be validated not just on signal fidelity, but on its capacity to preserve these subtle pathological signatures without introducing bias. To validate the clinical impact of our proposed method, we conducted a comprehensive evaluation using three distinct automated pipelines targeting different anatomical structures and modalities.

\subsubsection{Validation on Alzheimer's disease}

We selected a representative subset of 20 subjects from the Alzheimer's Disease Neuroimaging Initiative (ADNI) database curated from \cite{Coupe2017}. The cohort was balanced for demographics and diagnosis: 10 males ($72.12 \pm 8.70$ years) and 10 females ($75.08 \pm 6.39$ years), with an equal split of Cognitively Normal (CN, $N=10$) and Alzheimer’s Disease (AD, $N=10$) subjects. For each subject, we generated a comprehensive set of synthetic low-resolution degradations following the previously defined protocol at Section \ref{subsec:degradation}. To ensure statistical robustness across the degradation manifold, we generated samples for every anisotropy and volumetric cluster, resulting in a total of 420 evaluation volumes ($20 \text{ subjects} \times 7 \text{ bins} \times 3 \text{ anisotropies}$).

To assess global anatomical consistency and both global anatomical preservation and downstream diagnostic efficacy, we used AssemblyNet-AD \parencite{Nguyen2023TowardsIntelligence}, a validated deep learning pipeline designed for the automatic prediction of AD signatures from T1w MRI. The pipeline performs a sequence of preprocessing steps: denoising, MNI space affine registration, inhomogeneity correction, intensity normalization, intracranial cavity (ICC) extraction, and multi-structure segmentation (132 distinct regions). We integrated the resolution enhancement step immediately after denoising and prior to affine registration. This setup allows us to measure how resolution enhancement propagates through a complex neuroimaging pipeline.

\begin{figure}[h]

    \centering
    \includegraphics[width=\linewidth]{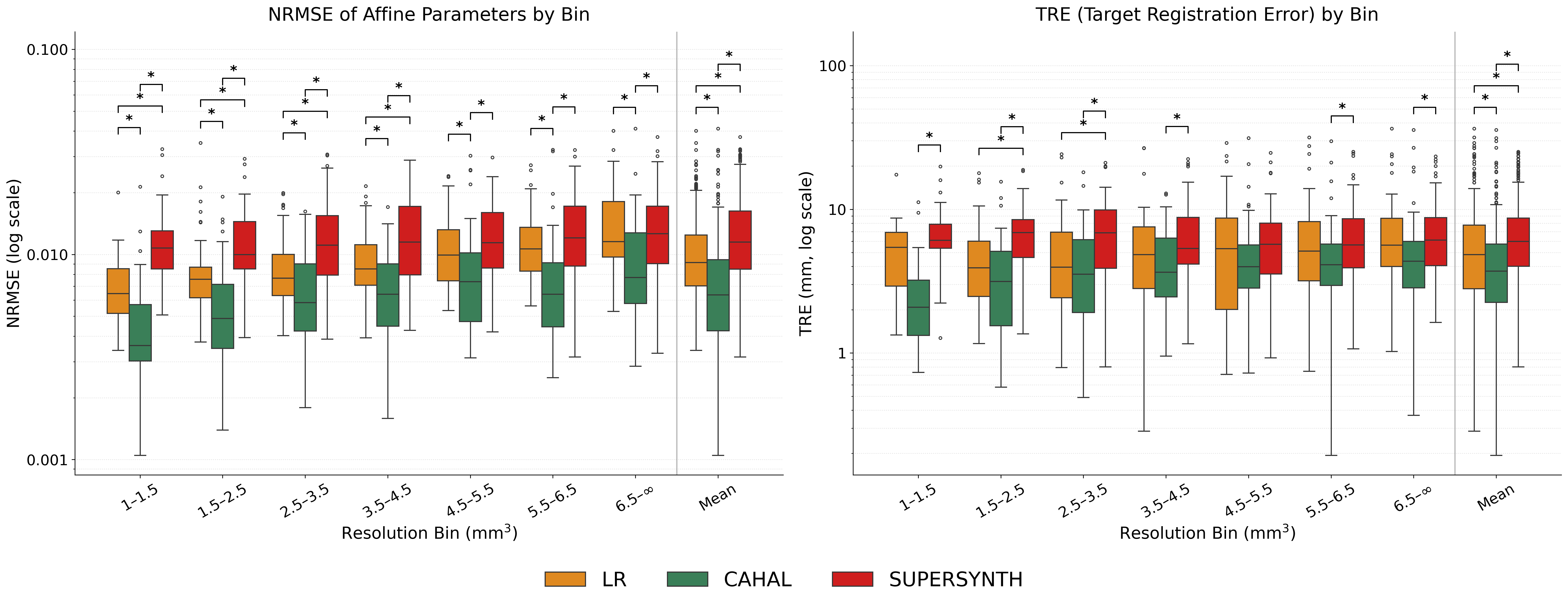}
    \caption{Impact of resolution improvement on automated registration. The proposed method was found to be significantly better than SuperSynth. $^*$ indicates statistical significance with p<0.01. Statistical test of TRE is computed in a linear scale, and a log scale is applied for visualisation. }
    \label{fig:reg_metrics}

\end{figure}

Initially, for segmentation pipelines operating within the MNI template space, downstream accuracy is intrinsically linked to the quality of the initial affine registration. We quantified this using the Target Registration Error (TRE) and the Normalized Root Mean Square Error (NRMSE) of the affine transformation parameters. As illustrated in Figure \ref{fig:reg_metrics}, image degradation significantly impedes registration accuracy. Notably, the generative baseline (SuperSynth) exhibits a higher NRMSE ($0.0179$) compared to the proposed CAHAL framework ($0.0104$) and even the standard LR baseline in certain regimes. This suggests that the stochastic "hallucinations" introduced by unconstrained generation can alter anatomical landmarks, leading the registration algorithm to converge on suboptimal local minima. In contrast, our residual approach consistently minimizes TRE across all anisotropy levels, ensuring that the restored geometry aligns faithfully with standard stereotaxic spaces.

\begin{figure}[htbp]
    \centering
    \includegraphics[width=0.875\linewidth]{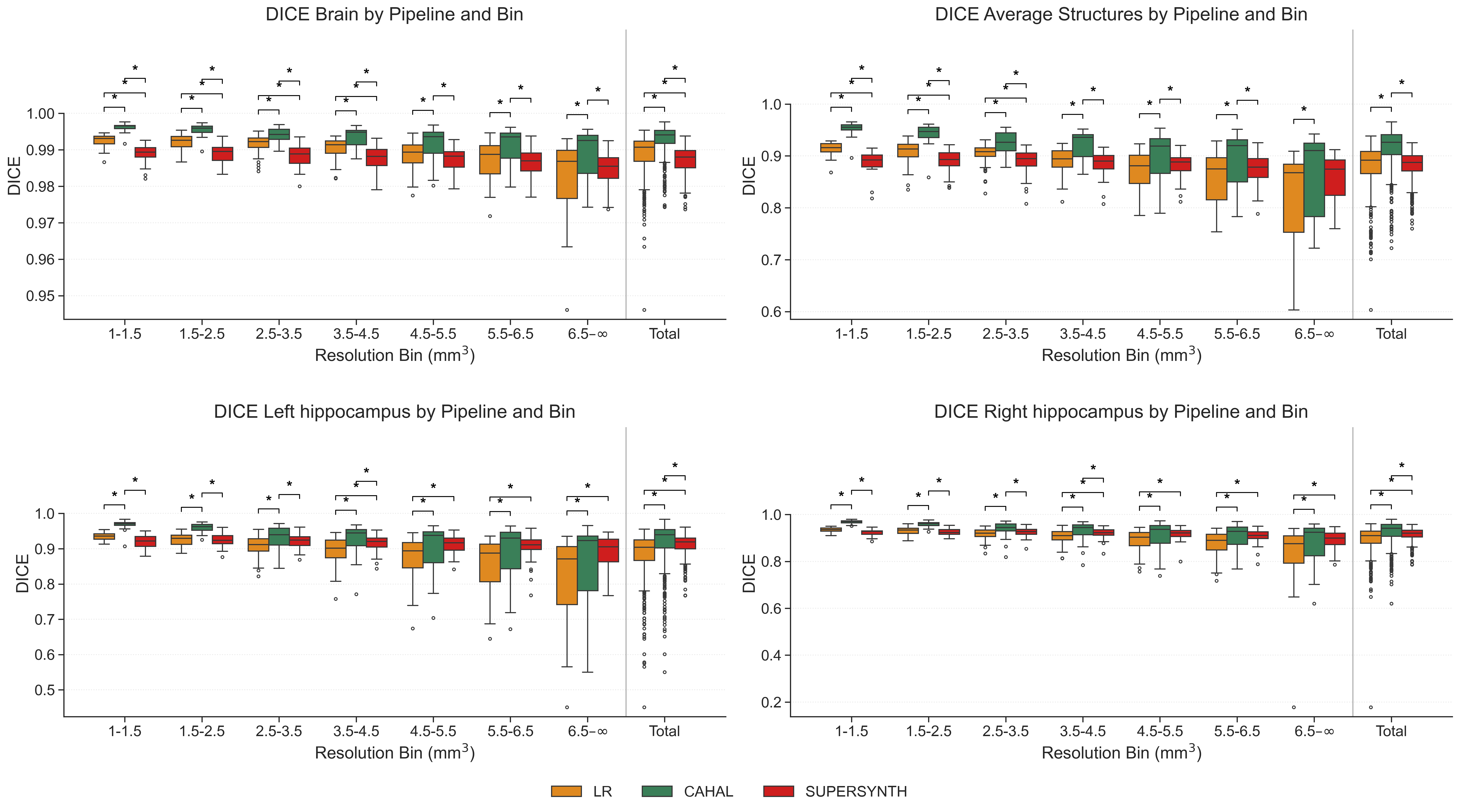}
        \caption{Preservation of structural segmentation accuracy. Subplots display the Dice for: whole-brain segmentation (top left), the macro-average of all 132 anatomical structures (top right), the left hippocampus (bottom left), and the right hippocampus (bottom right). CAHAL obtained the best results at all resolution bins and labels. $^*$ indicates statistical significance with p<0.01.}
        \label{fig:seg_assemblynet}
    \end{figure}

Next, we evaluated the Dice Similarity Coefficient (DSC) for the Intracranial Cavity (ICC) and key subcortical structures. Of particular interest is the Hippocampus, a primary biomarker for AD atrophy \parencite{Frisoni2017StrategicBiomarkers}. Figure \ref{fig:seg_assemblynet} demonstrates that while standard interpolation (LR) leads to a rapid decline in segmentation accuracy as slice thickness increases (Bins 6-7), CAHAL maintains high Dice scores ($>0.90$), effectively mitigating the degradation of the segmentation network caused by resolution loss. Crucially, while SuperSynth performs adequately in high-resolution bins, it exhibits drop-offs in the most degraded regimes. This confirms that our physics-informed MoE experts are essential for preserving the distinct boundaries of small, complex structures like the hippocampus against partial volume effects.

\begin{figure}[H]
    \centering
    \includegraphics[width=0.8\linewidth]{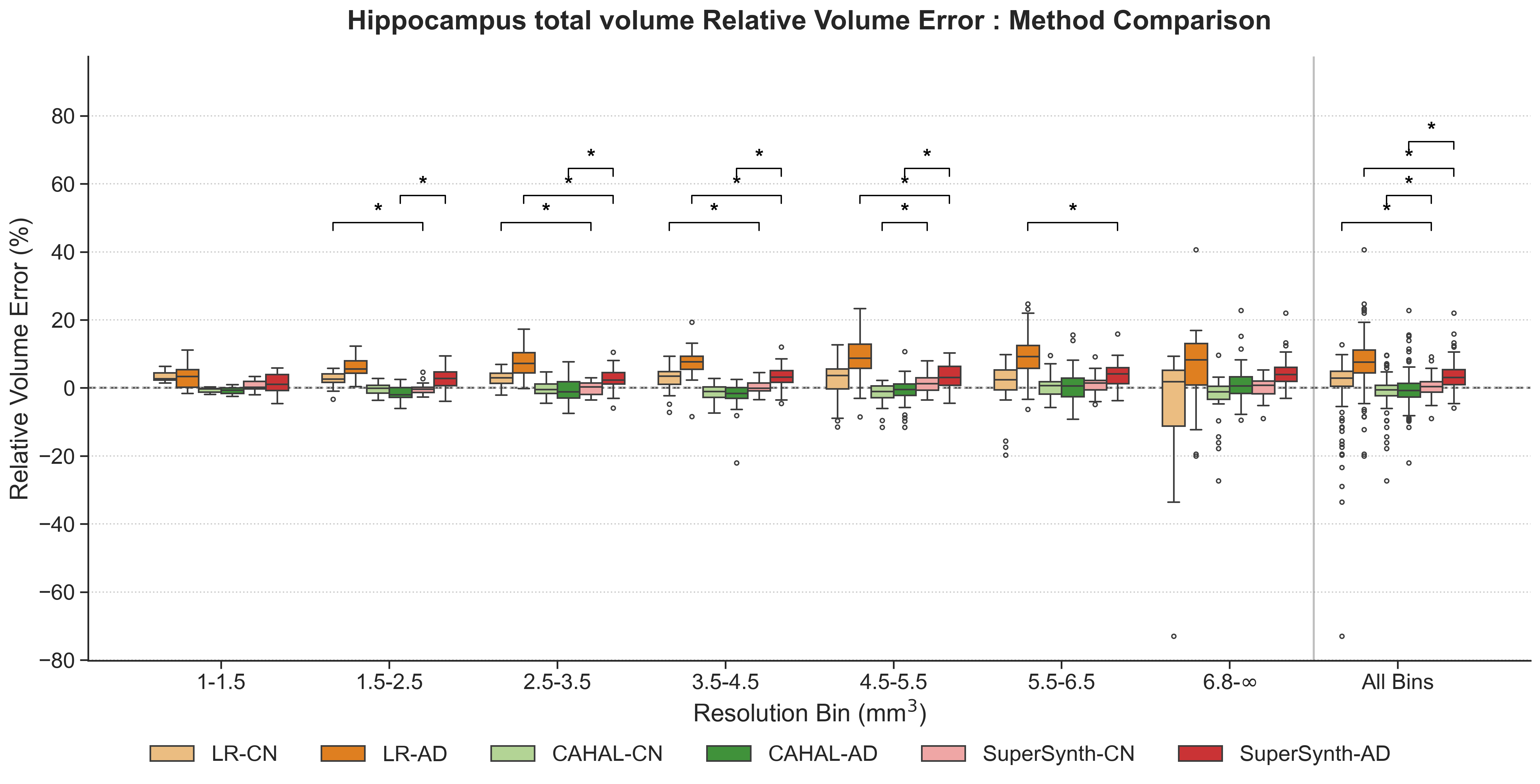}
        \caption{Generative volumetric bias. Statistical significance between method pairs within each diagnosis group (CN or AD) is indicated. $^*$ indicates statistical significance with p<0.01.}
        \label{fig:hippocampues_rve}
    \end{figure}

\begin{figure}[H]
    \centering
    \includegraphics[width=\linewidth]{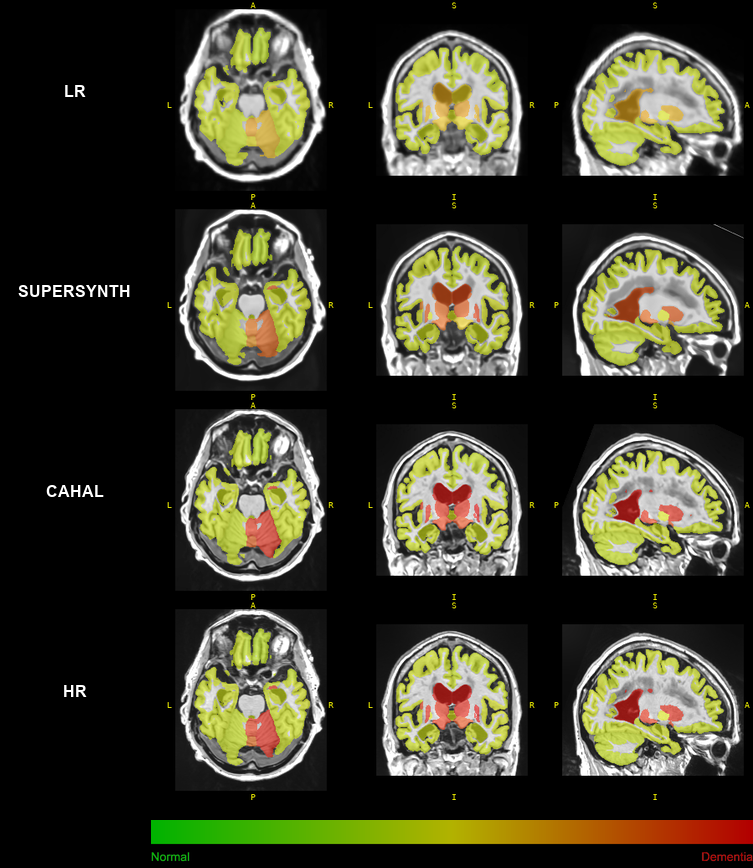}
    \caption{Impact of resolution enhancement on automated disease grading spatial maps (DeepGrading score) \cite{Nguyen2023DeepDementia} for an AD case of $1.5 \times 2.6 \times 1.5 \text{ mm}^3$. Note that CAHAL recovers better the grading score information than SuperSynth, driving more similar patterns compared to the high resolution image.}
    \label{fig:assemblynet_ad_case}
\end{figure}

Related to these results, we found recent literature cautions against using label-based deep learning-based synthetic MRI (simulated) for volumetry. For instance, \parencite{Riedel2026DeepDiseases} has recently demonstrated that generative models (such as the SynthSR family,in this case SynthSR) introduce systematic errors by consistently overestimating regional and global brain volumes, thereby failing to preserve the true atrophy patterns required for accurate automated disease prediction. Furthermore, established volumetric pipelines (such as those sharing foundations with FreeSurfer) exhibit their own well-documented internal biases, specifically, the artificial overestimation of large hippocampus and the underestimation of small, atrophic hippocampus \parencite{Piot2025EstimationEffects, Brookmeyer2017AMethods}. We directly corroborated these limitations within our experimental setup. To evaluate structural bias, we analyzed the relative volume error (RVE), which measures the percentage deviation of the predicted volume from the high-resolution ground truth. As illustrated in Figure \ref{fig:hippocampues_rve}, this analysis reveals that the generative SuperSynth baseline consistently overestimates total hippocampal volume across all spatial resolution bins. This systemic inflation is further corroborated by the Bland-Altman plots ( see Figure \ref{fig:grading_blandaltaman} in Appendix). This generative volume inflation significantly impacts the Alzheimer's Disease (AD) cohort by artificially enlarging atrophic hippocampi. As a result, it masks the true morphological signature of the disease. Conversely, our proposed CAHAL framework maintains an RVE tightly centered at zero for both Cognitively Normal (CN) and AD subjects across all degradation regimes. This demonstrates that CAHAL faithfully preserves true anatomical dimensions, preventing hallucinated tissue inflation and maintaining the structural separation necessary for reliable biomarker extraction. A qualitative example of the grading score for the degraded LR, the original HR and the restored volume with the two methods can be seen in Figure \ref{fig:assemblynet_ad_case} at Appendix.

These compounding biases can artificially inflate or deflate population separation regardless of the actual image fidelity. To ensure a fair and rigorous evaluation, we focus our primary analysis on the grading metric \parencite{Nguyen2023DeepDementia} from AssemblyNet-Ad. The classification pipeline of this framework operates in two distinct stages. First, in the classification part of this pipeline, it creates the Deep Grading Map, which is a 3D spatial representation that assigns an individual AD-similarity score to each segmented anatomical structure. This map serves as a crucial tool for clinical interpretability, as you can see in Figure \ref{fig:assemblynet_ad_case}. Second, these localized features are processed by a Graph Convolutional Network (GCN) to capture the complex topological relationships between brain structures, an approach demonstrated by the original authors to be highly effective scoring mechanism relies on intrinsic textural and structural patterns rather than susceptible volumetric boundaries or hallucinated morphologies, it serves as a highly sensitive probe for true pathological signal preservation. The final grading score is a continuous probability ranging from 0 (Cognitively Normal) to 1 (Alzheimer's Disease). To prevent the introduction of confounding variables from secondary variables, we computed the Receiver Operating characteristic Curve (ROC), the Area Under the Curve (AUC) and Intraclass Correlation Coefficient (ICC) directly using these raw continuous scores.

\begin{figure}[h]
    \centering
    % First row
    \begin{subfigure}{\textwidth}
        \centering
        \includegraphics[width=0.8\linewidth]{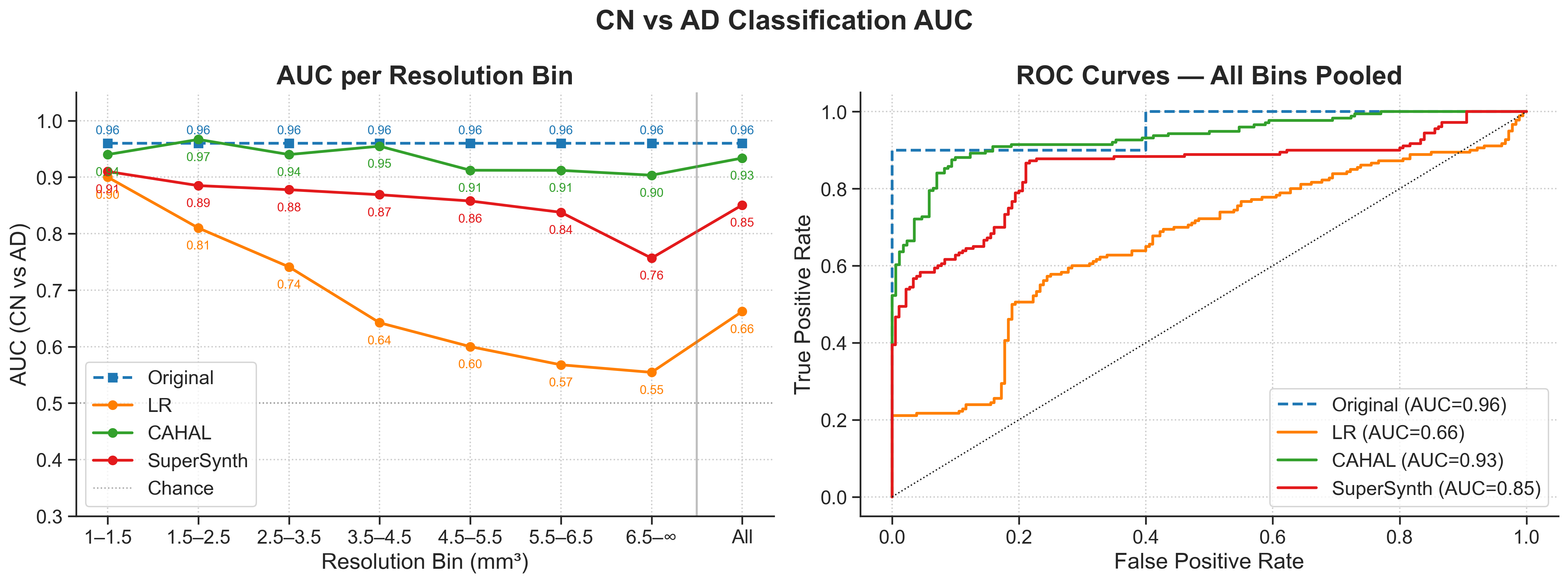}
        \caption{Classification performance (Cognitively Normal vs. Alzheimer's Disease) based on the grading score.}
        \label{fig:grading_auc}
    \end{subfigure}

    \vspace{0.5cm}
    
    \begin{subfigure}{\textwidth}
        \centering
        \includegraphics[width=0.8\linewidth]{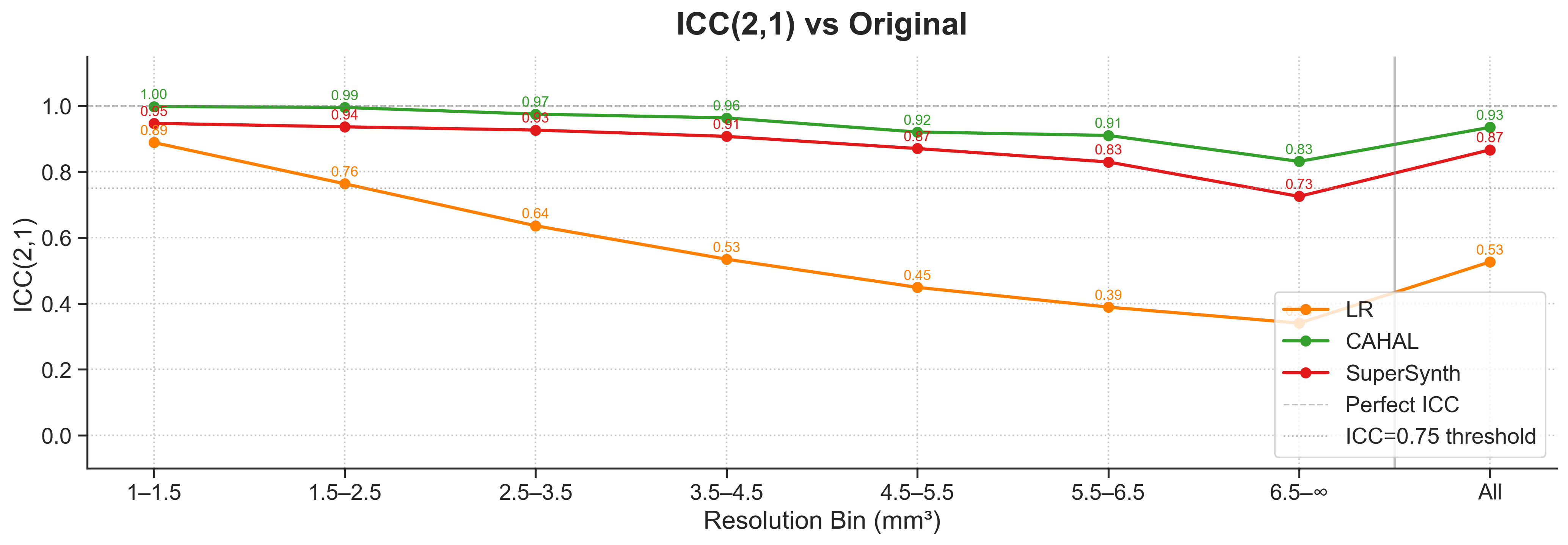}
        \caption{Intraclass Correlation Coefficient (ICC) measuring agreement with the original High-Resolution grading scores.}
        \label{fig:grading_icc}
    \end{subfigure}

    \caption{Preservation of diagnostic power and signal fidelity.}
    \label{fig:ad_metrics}
\end{figure}

The primary objective of a clinical SR tool is to recover diagnostic information lost during low-resolution acquisition. As demonstrated in Figure \ref{fig:grading_auc} (ROC Curves, all bins pooled), the original High-Resolution (HR) scans possess an AUC of $0.96$ for CN vs. AD classification. When analyzing the degraded Low-Resolution (LR) inputs, this discriminative power drops to an AUC of $0.66$, effectively rendering the scans non-diagnostic. CAHAL successfully restores this pathological signal, achieving a pooled AUC of $0.93$. Crucially, as seen in the bin-wise AUC analysis, CAHAL maintains this high diagnostic accuracy ($>0.90$) consistently across all degradation severities. In contrast, the generative SuperSynth baseline only recovers an AUC of $0.85$, with its performance deteriorating significantly ($<0.80$) in highly degraded regimes (Bins 6-7). The discrepancy in diagnostic power between CAHAL and SuperSynth highlights a critical risk of generative architectures in clinical workflows. While generative models excel at synthesizing visually pleasing, brain-like textures, these unconstrained "hallucinations" often overwrite subtle, disease-specific textural signatures required for grading. This structural drift is quantitatively confirmed in our agreement analysis against the original HR grading scores. In the Intraclass Correlation Coefficient (ICC) analysis (Figure \ref{fig:grading_icc}), CAHAL maintains near-perfect agreement (ICC $\approx 0.96-0.99$) across all resolution bins. Conversely, SuperSynth introduces substantial variance and unpredictable alterations to the grading score (detailed Bland-Altman limits of agreement are provided in the Appendix, Figure \ref{fig:grading_blandaltaman}).

\FloatBarrier

\subsubsection{Validation on ultra-high resolution structures }

To test the proposed method's limit on intricate anatomical details, we also employed DeepCERES \parencite{Morell-Ortega2025DeepCERES:MRI}, a pipeline dedicated to cerebellar lobule segmentation. The cerebellum features extremely thin cortical foliations and complex white matter branches (Arbor Vitae) that are easily obliterated by partial volume effects. We focused on the segmentation of Cerebellar White Matter, a structure requiring high-frequency detail preservation. 

\begin{figure}[h!]
    \centering
    \includegraphics[width=\linewidth]{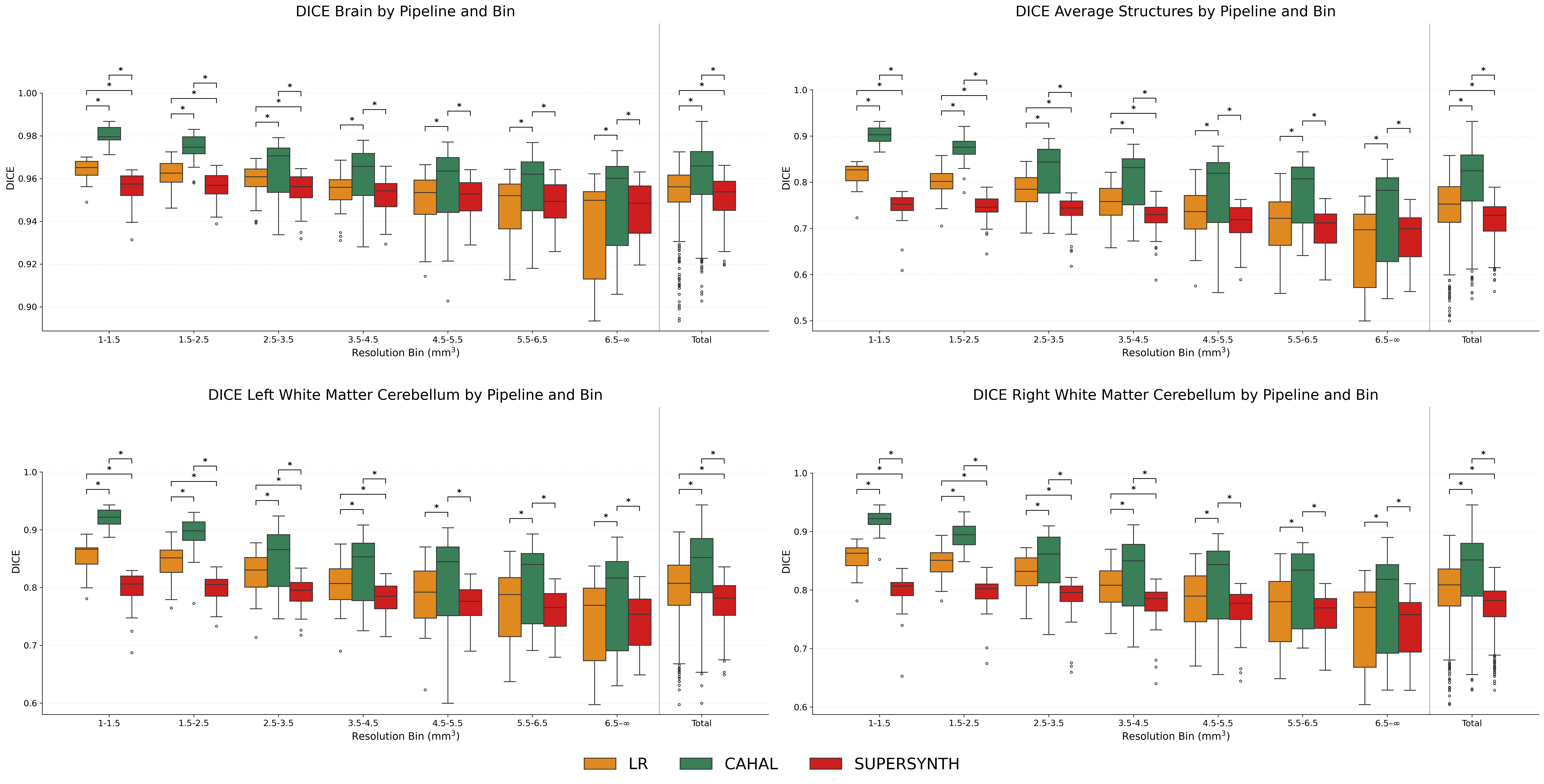}
    \caption{Validation on ultra-high resolution cerebellar structures. Subplots display the Dice for: whole-brain segmentation (top left), the macro-average of all 27 cerebellum substructures (top right), the left cerebellar white matter (bottom left), and the right cerebellar white matter (bottom right). As you can see, CAHAL obtained the best results at all resolution bins and labels. $^*$ indicates statistical significance with p<0.01.}
    \label{fig:deep_ceres_dice}
\end{figure}

As shown in Figure \ref{fig:deep_ceres_dice}, our method demonstrates superior retention of these fine structures compared to generative baselines. Notably, the quantitative results reveal that SuperSynth frequently underperforms the unenhanced Low-Resolution (LR) input across most volumetric degradation bins. This decrease in segmentation accuracy indicates that unconstrained generative synthesis actively merges or erodes delicate high-frequency boundaries rather than restoring them. In contrast, CAHAL robustly improves segmentation accuracy over the LR baseline and maintains stable performance across the resolution manifold.

\begin{figure}[H]
    \centering
    \includegraphics[width=\linewidth]{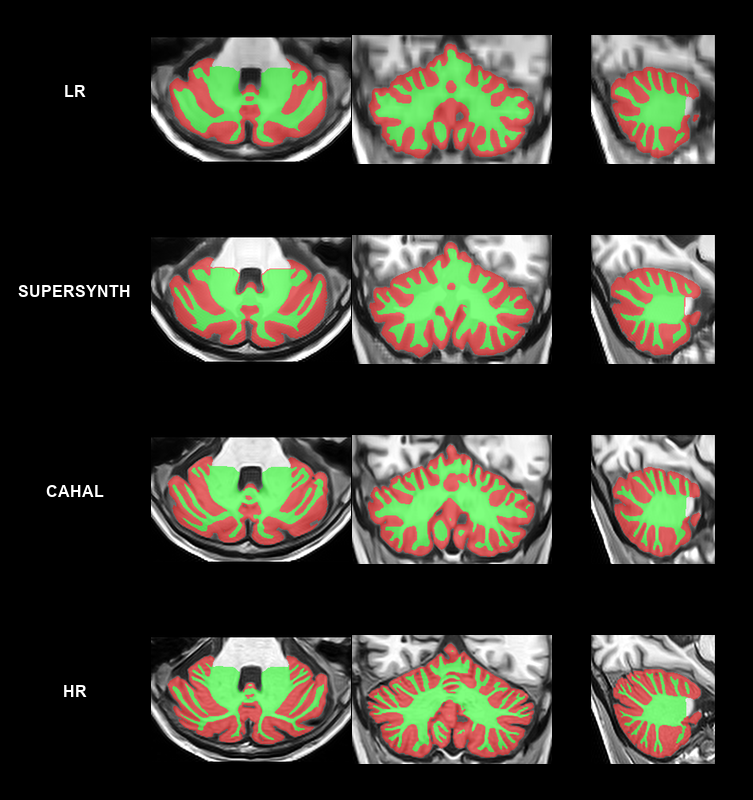}
    \caption{Qualitative evaluation of a moderately anisotropic clinical acquisition ($1.3 \times 1.3 \times 3.3 \text{ mm}^3$) with DeepCERES pipeline. The white matter is represented in green, while the grey matter is shown in red. Note that CAHAL better recovers the cerebellum white matter than SuperSynth.}
    \label{fig:deepceres_example}
\end{figure}

This quantitative trend is supported by our qualitative assessments (Appendix Figure \ref{fig:deepceres_example}). By explicitly modeling the point-spread function via our bivariate expert grid, CAHAL accurately recovers the fine white matter digitations that SuperSynth tends to artificially over-smooth into a simplified mass, further supporting the necessity of anisotropy-aware model selection for fine-grained neuroanatomy.

\subsubsection{Validation on pathological lesions (FLAIR)}

Finally, to quantify the diagnostic utility of our method on FLAIR acquisitions, we evaluated its impact on a downstream lesion segmentation task. Specifically, we utilized DeepLesionBrain \parencite{Kamraoui2022DeepLesionBrain:Segmentation}, a specialized pipeline for Multiple Sclerosis (MS) lesion detection, to quantify how our resolution enhancement affects the reliability of automated white matter lesion segmentation. For this evaluation, we selected five independent cases from the volBrain database, ensuring strict exclusion from the training distribution to prevent data leakage. We applied the identical set of synthetic degradations used in the AD experiment to these subjects (resulting in 5x7x3 degraded images). Importantly, to isolate the impact of the FLAIR reconstruction, we restricted these degradations exclusively to the FLAIR volumes while retaining the paired T1-weighted (T1w) images at their original high resolution.

\begin{figure}[H]
    \centering
    \includegraphics[width=\linewidth]{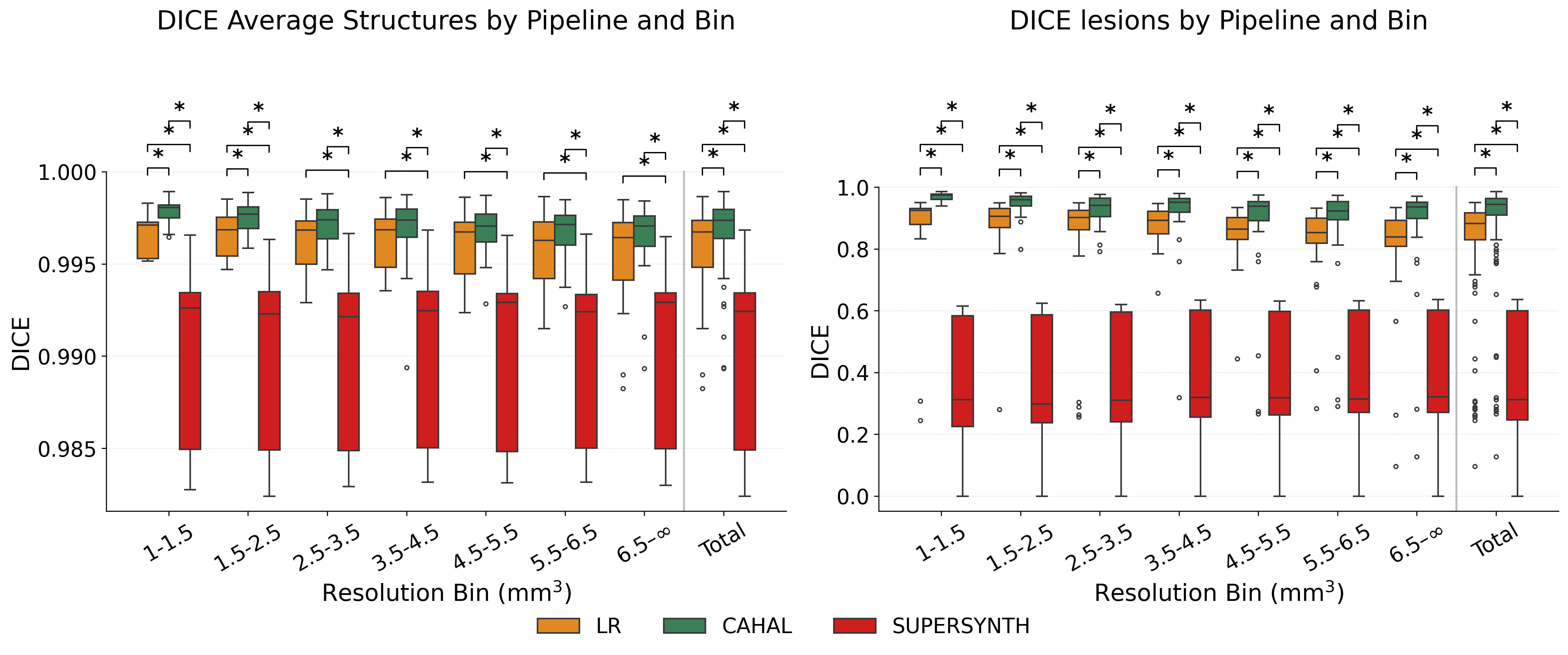}
    \caption{Validation on white matter lesion segmentation. Left: average Dice of 132 brain structures. Right: average DICE of lesion mask. $^*$ indicates statistical significance with p<0.01.}
    \label{fig:lesion_dice}
\end{figure}

DeepLesionBrain relies heavily on the spatial coherence between the high-resolution T1 and FLAIR inputs to accurately delineate tissues (the segmentation networks take as input both modalities). As shown in the dice of structures at Figure \ref{fig:lesion_dice} (left), the generative approach exhibits a critical failure mode here. SuperSynth introduces unconstrained anatomical modifications that effectively decouple the spatial alignment between the T1 and FLAIR volumes. This loss of inter-modality coherence severely disrupts the segmentation network, resulting in a notable drop in the average structural Dice score compared to the original inputs.

Beyond global structural drift, the generative model actively distorts the pathology itself. Figure \ref{fig:lesion_dice} (right) presents the lesion segmentation performance, revealing that SuperSynth suffers from a severe drop in Lesion Dice, performing worse than the unenhanced Low-Resolution (LR) input. This confirms the severe clinical risk of "hallucination bias" (or pathological inpainting): generative models, driven to synthesize a plausible image from a learned distribution, tend to regress towards a healthy mean. Consequently, they either reconstruct true MS lesions as normal white matter (erasing pathology) or generate noise artifacts that mimic lesions (creating false positives).This effect is visually demonstrated in Appendix Figure \ref{fig:deeplesion_brain_example1} and Figure \ref{fig:deeplesion_brain_example2}, where the generative baseline fails to delineate several periventricular lesions present in the ground truth, effectively 'inpainting' them with healthy tissue. Conversely, CAHAL maintains lesion detection accuracy (Dice $> 0.90$) even in the lowest resolution bins, 
and preserves the inter-modality spatial coherence between T1w and FLAIR volumes that the segmentation pipeline requires. The faithful preservation of lesion 
boundaries and locations by CAHAL has direct implications for advanced downstream analyses dependent on accurate lesion maps, as discussed in Section~\ref{sec:discussion}.
% Conversely, CAHAL maintains lesion detection accuracy (Dice > 0.90) even in the lowest resolution bins. By strictly adhering to a residual formulation, our model enhances the contrast and sharpness of hyperintensities without altering their underlying topology or disrupting the cross-modality anatomical alignment. This ensures that the true diagnostic sensitivity of the scan is faithfully preserved, proving the safety of the residual MoE approach for clinical pipelines.

% The clinical implications of this pathological distortion extend far beyond simple volumetry, they fundamentally compromise advanced downstream network analyses. In MS research, accurate lesion maps are frequently projected onto white matter tractography atlases to compute the "disconnectome" \parencite{Ravano2021,ThiebautdeSchotten2020}, a quantitative map of the specific structural neural networks severed by the disease. Because generative baselines like SuperSynth hallucinate healthy tissue over true lesions, the computed disconnectome will falsely predict intact neural pathways, dangerously underestimating the patient's true neurological burden. By better preserving the boundaries and locations of pathological hyperintensities, the CAHAL approach facilitates a more reliable computation of the patient's derived disconnectome. This demonstrates the clinical safety advantage of employing residual-based SR frameworks within automated clinical pipelines.

\section{Discussion}
\label{sec:discussion}

The results presented across Sections~\ref{sec:comparison} 
and~\ref{sec:validation} reveal a consistent and clinically 
meaningful divide between the two methodological paradigms 
evaluated. Generative methods optimise for distributional 
plausibility, synthesising high-frequency detail that 
matches the appearance of a training label distribution 
rather than the patient's true anatomy. This produces 
visually compelling outputs but introduces systematic 
deviations detectable across every metric: reduced PSNR, 
degraded CC, elevated LPIPS, and impaired segmentation 
accuracy. The LPIPS analysis is particularly instructive: 
SuperSynth achieves perceptual plausibility comparable to 
the low-resolution reference while falling below it on 
every structural metric, demonstrating that an image can 
appear anatomically plausible without faithfully 
representing the underlying patient anatomy, and that 
perceptual metrics alone are insufficient proxies for 
clinical safety. Residual methods constrain the network 
output to remain anchored to the original native signal, 
making the architecture structurally incapable of 
replacing anatomical structures with synthesised 
alternatives, a property with direct clinical 
implications wherever pathology is present.

The comparison with REMIX provides the most direct 
empirical evidence for the clinical necessity of 
native-space processing. Despite sharing the residual 
reconstruction paradigm, REMIX suffers a segmentation 
accuracy degradation that grows monotonically with 
acquisition anisotropy, from a marginal deficit at 
near-isotropic resolutions to a substantial gap at 
Bin 6 (Dice: 0.799 vs. reference 0.863). This gradient 
constitutes quantitative evidence that MNI-space 
round-trip interpolation introduces geometric 
distortions whose severity scales with through-plane 
anisotropy. Together with the registration error 
analysis, which shows that SuperSynth's hallucinated 
boundaries actively mislead registration, while CAHAL 
provides accurate anatomical landmarks, these findings 
validate native-space processing not as a convenience 
but as a clinical safety requirement for SR integration 
into automated morphometric pipelines.

The clinical validation results demonstrate that these 
signal-level advantages translate directly into 
diagnostic utility. In Alzheimer's Disease, CAHAL 
recovered the AD classification AUC from a 
non-diagnostic 0.66 to 0.93, approaching the 
high-resolution ceiling of 0.96, while SuperSynth 
recovered only 0.85 and deteriorated below 0.80 in 
the most degraded regimes. Critically, SuperSynth 
systematically inflated hippocampal volumes,
artificially enlarging atrophic tissue and thereby 
masking the morphological signature of the disease,  
while CAHAL maintained a relative volume error tightly 
centred at zero across all degradation bins and 
diagnostic groups. This elimination of volumetric bias 
is a prerequisite for reliable biomarker extraction 
that generative methods cannot currently provide. In 
the FLAIR evaluation, generative synthesis exhibited 
pathological inpainting, erasing true MS lesions or 
hallucinating false ones, and disrupting the 
cross-modality spatial coherence that lesion 
segmentation pipelines depend on. CAHAL maintained 
lesion detection accuracy above Dice 0.90 even in the 
lowest resolution bins, and preserved the inter-modality 
alignment required for accurate disconnectome 
computation, a downstream analysis whose validity is 
fundamentally compromised by any hallucination of 
lesion boundaries.

Beyond native-space processing, the bivariate expert 
partitioning provides a second independent source of 
benefit. Volumetric resolution alone is insufficient 
as a routing criterion: a volume-only MoE fails in 
highly anisotropic regimes where a single thick-slice 
axis presents a fundamentally different restoration 
problem from isotropic blur at the same voxel volume. 
Conditioning expert selection on both V and A recovers 
an additional 2.0 dB over the volume-only baseline, 
with the largest gains in the highest-anisotropy bin, 
while the deterministic gating ensures this 
architectural complexity incurs no inference overhead.

Several limitations and future directions warrant 
acknowledgement. First, all quantitative evaluations 
use synthetically degraded inputs from the same 
pipeline used for training, which may artificially 
favour CAHAL relative to natively acquired clinical 
pairs. Validation on paired native LR/HR acquisitions 
from a held-out clinical site is the most important 
outstanding requirement for full clinical translation. 
Second, sequential curriculum training incurs a 
linearly scaling computational cost; future work will 
explore dynamic parameter sharing and sparse routing 
to reduce this footprint while preserving 
specialisation. Third, while validated on T1-weighted 
and FLAIR sequences, extension to T2-weighted and 
diffusion-weighted imaging represents a natural next 
step, likely requiring dedicated modality-specific 
fine-tuning. As a preliminary investigation, we 
evaluated the T1w-trained experts on FLAIR data 
without fine-tuning (Table~\ref{tab:flair_psnr_crossmod}, 
Appendix). While the dedicated FLAIR model 
substantially outperforms the T1w transfer, the T1w 
experts nonetheless achieve non-trivial FLAIR 
restoration, suggesting that the residual formulation 
learns anatomical representations that partially 
generalise across contrasts. This motivates the 
longer-term development of a foundational multi-modal 
model trained jointly across contrasts, one that, 
unlike generative foundational approaches such as 
SynthSR and SuperSynth, would maintain structural 
fidelity as a hard architectural constraint. Extending 
the bivariate MoE with a modality conditioning axis 
represents a promising direction toward a clinically 
safe, protocol-agnostic enhancement framework.

\section{Conclusion}
\label{sec:conclusion}

CAHAL demonstrates that clinically safe brain MRI 
resolution enhancement requires faithful residual 
reconstruction over unconstrained generation. By 
combining physics-informed degradation modelling, 
bivariate expert specialisation, and native-space 
processing, CAHAL consistently outperforms 
state-of-the-art generative baselines across signal 
fidelity, structural preservation, and downstream 
clinical utility. It is the only method evaluated 
that both exceeds the low-resolution reference on 
segmentation accuracy and restores near-ground-truth 
diagnostic power across the full spectrum of clinical 
degradation profiles, while eliminating the 
volumetric bias that undermines automated biomarker 
extraction in neurodegenerative disease. 

% By making 
% code and pretrained models publicly available and 
% integrating CAHAL into the volBrain platform, we 
% aim to facilitate the reliable use of large-scale 
% retrospective clinical MRI databases for automated 
% neuroimaging research.

\section{Data and Code Availability}
To promote open science, ensure full methodological reproducibility, and facilitate the immediate adoption of safe, high-fidelity super-resolution in clinical research, the CAHAL framework will be open-sourced. The complete inference pipeline, including the deterministic bivariate routing mechanism and the fully trained Mixture of Experts (MoE) weights for both T1-weighted and FLAIR modalities, will be made freely available to the research community at \href{https://github.com/volBrain-net/CAHAL}{https://github.com/volBrain-net/CAHAL}. Furthermore, an implementation of CAHAL will be integrated into the automated preprocessing pipelines of the public volBrain platform (volbrain.net). This will allow clinicians and researchers worldwide to seamlessly enhance their native MRI data without the need for local GPU infrastructure.

\section*{Acknowledgments}
This work has been developed thanks to the project PID2023-152127OB-I00 of the Ministerio de Ciencia e Innovacion de España. We thank the support of ITACA (Institute of Information and Communication Technologies) at UPV (Universitat Politècnica de València). This work also benefited from the support of the project DeepvolBrain and HoliBrain of the French National Research Agency (ANR-18-CE45-0013 and ANR-23-CE45-0020-01). This study was achieved within the Laboratory of Excellence TRAIL ANR-10-LABX-57 for the BigDataBrain project. Moreover, we thank the Investments for the future Program IdEx Bordeaux (ANR-10- IDEX- 03- 02, HL-MRI Project), Cluster of excellence CPU and the CNRS.

\newpage

\printbibliography

\newpage
\section*{Appendix}

\begin{figure}[htbp]
    \centering
    % First Subfigure: T1w
    \begin{subfigure}[b]{0.48\textwidth}
        \centering
        \includegraphics[width=\textwidth]{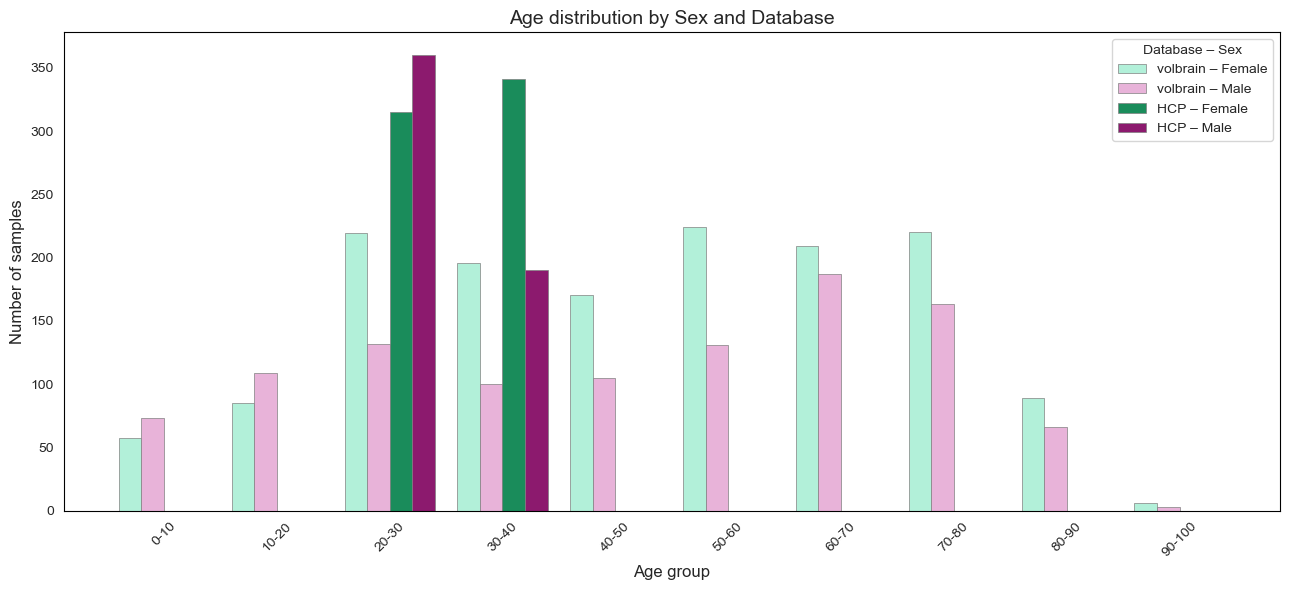}
        \caption{T1w Age Distribution ($N_{VOLBRAIN}=3,715 + N_{HCP}=1,027$)}
        \label{fig:age_t1}
    \end{subfigure}
    \hfill % Adds horizontal space between the two images
    % Second Subfigure: FLAIR
    \begin{subfigure}[b]{0.48\textwidth}
        \centering
        \includegraphics[width=\textwidth]{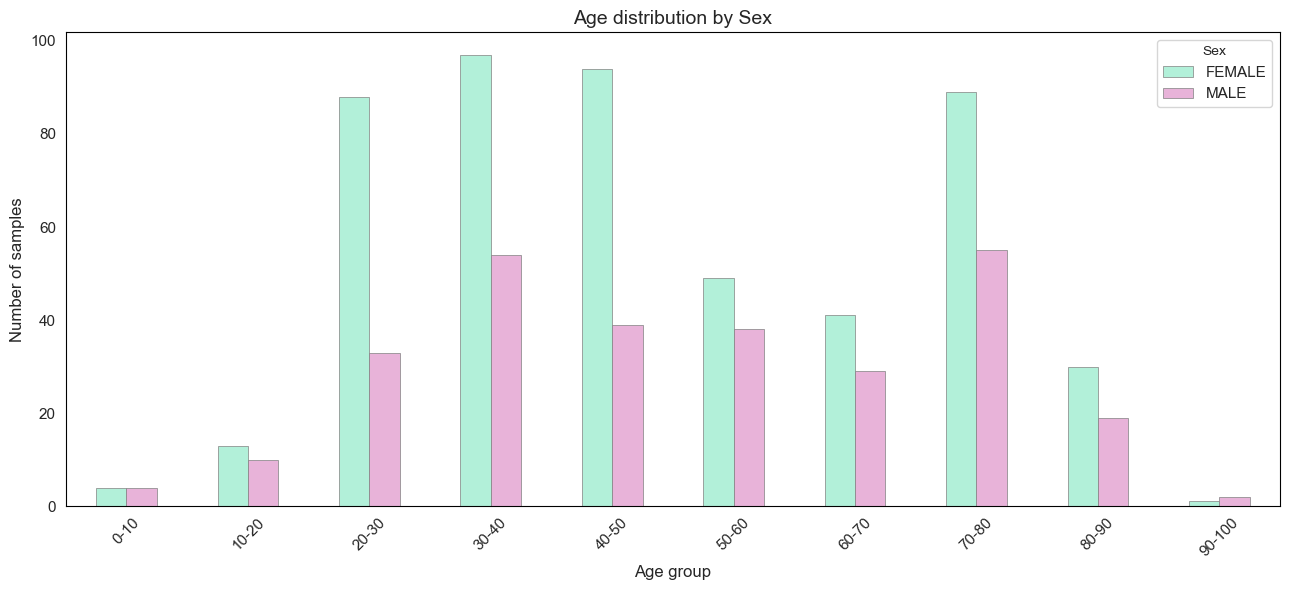}
        \caption{FLAIR Age Distribution ($N=1,975$)}
        \label{fig:age_flair}
    \end{subfigure}
    
    \caption{\textbf{Demographic profile of the curated training cohorts.} Histograms illustrate the age distribution hued by biological sex. The combined datasets provide comprehensive coverage across the human lifespan, ensuring the model's exposure to age-related structural variance and diverse brain morphologies.}
    \label{fig:demographics_histograms}
\end{figure}

\begin{table}[htbp]
\centering
\caption{\textbf{Geographic Distribution of Training Data.} Values represent the subset of cases with available metadata. Percentages are calculated relative to the total number of known samples per modality.}
\label{tab:geographic_distribution}
\begin{tabularx}{\textwidth}{X c c c c}
\toprule
 & \multicolumn{2}{c}{\textbf{T1w}} & \multicolumn{2}{c}{\textbf{FLAIR}} \\
\cmidrule(lr){2-3} \cmidrule(lr){4-5}
\textbf{Continent} & \textbf{N} & \textbf{\%} & \textbf{N} & \textbf{\%} \\
\midrule
Europe             & 1,034 & 40.6\% & 152 & 19.3\% \\
Asia               & 620   & 24.4\% & 133 & 16.9\% \\
South America      & 385   & 15.1\% & 28  & 3.5\%  \\
Central America    & 275   & 10.8\% & 224 & 28.4\% \\
North America      & 147   & 5.8\%  & 232 & 29.4\% \\
Oceania            & 43    & 1.7\%  & 0   & 0.0\%  \\
Africa             & 40    & 1.6\%  & 20  & 2.5\%  \\
\midrule
\textbf{Total (Known)} & \textbf{2,544} & \textbf{100\%} & \textbf{789} & \textbf{100\%} \\
\bottomrule
\end{tabularx}
\end{table}

% \begin{figure}[htpb]
%     \centering
    
%     % Top Figure: Pearson Correlation Coefficient
%     \includegraphics[width=0.9\textwidth]{cc_boxplot.png}
    
%     \vspace{0.5cm} % Adds a clean gap between the two images
    
%     % Bottom Figure: Inference Time
%     \includegraphics[width=0.9\textwidth]{time_boxplot_split.png}
    
%     \caption{Evaluation of hardware adaptability and computational efficiency. The benchmark was conducted on 35 test volumes (5 cases sampled from each of the 7 spatial resolution bins) across six computational tiers. The x-axis denotes the hardware tier and corresponding minimum patch size used. \textbf{(Top) Pearson Correlation Coefficient (CC):} Structural fidelity remains stable ($>0.99$) across all memory tiers, indicating that dynamic tiling maintains precision without introducing significant boundary artifacts. \textbf{(Bottom) Inference Time:} Processing time in seconds (broken y-axis scale accommodates CPU execution). While increased tiling on smaller GPUs introduces a temporal overhead, inference remains under 10 seconds across all evaluated GPU configurations.}
%     \label{fig:hardware_benchmark}
% \end{figure}

\begin{figure}[H]
    \centering
    \includegraphics[width=\linewidth]{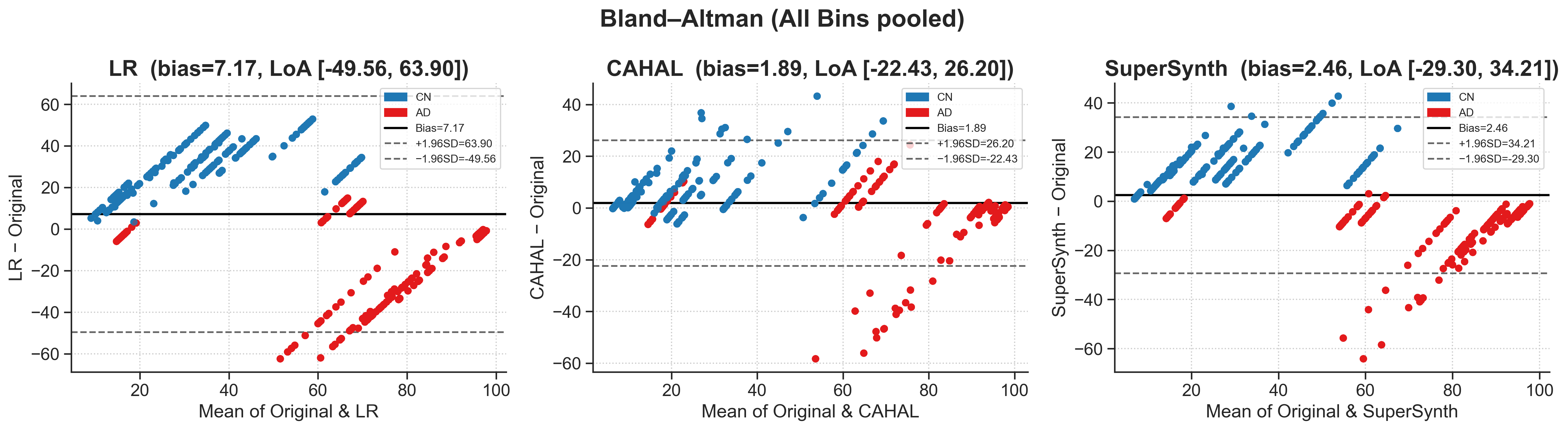}
    \caption{Bland-Altman analysis of diagnostic signal agreement (grading score). Comparison of native LR, our CAHAL framework, and SuperSynth against the ground truth for Control (blue) and Alzheimer's (red) subjects. The horizontal axis represents the mean of the two measurements, while the vertical axis plots their difference (Prediction $-$ Original). Solid lines indicate the mean systemic bias. Dashed lines represent the 95\% Limits of Agreement (LoA), defining the expected error margin for 95\% of the predictions. CAHAL demonstrates the highest fidelity, achieving the lowest systemic bias (1.89) and tightest LoA. }
    \label{fig:grading_blandaltaman}
\end{figure}

\begin{figure}[H]
    \centering
    \includegraphics[width=\linewidth]{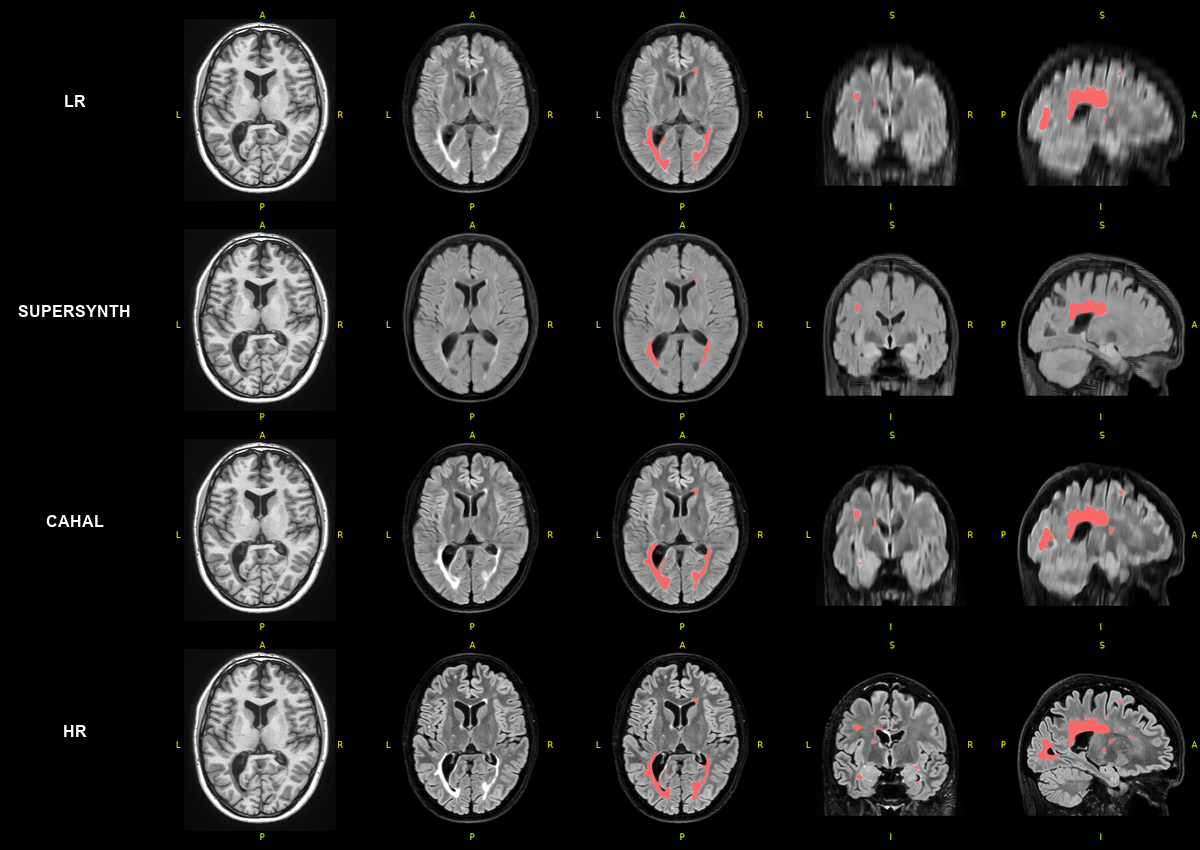}
    \caption{Qualitative reconstruction of severe through-plane anisotropic degradation ($1 \times 1 \times 8.5 \text{ mm}^3$). Lesion masks in red. Note that CAHAL recovers better the MS lesion boundaries than SuperSynth, driving more similar segmentation patterns compared to the high-resolution image.}
    \label{fig:deeplesion_brain_example1}
\end{figure}

\begin{figure}[H]
    \centering
    \includegraphics[width=\linewidth]{images/lesion_segmentation_comparison_res1.73_1.73_3.01_job377964t1.png}
    \caption{Preservation of pathological segmentation boundaries of a  $1.73 \times 1.73 \times 3.01 \text{ mm}^3$ case. Lesion masks in red. Note that CAHAL recovers better the pathological hyperintensities than SuperSynth, driving more similar MS lesion patterns compared to the high-resolution image.}
    \label{fig:deeplesion_brain_example2}
\end{figure}

\begin{table}[H]
\centering
\caption{\textbf{FLAIR modality — PSNR (mean $\pm$ standard deviation).} 
\textbf{Bold} indicates best performance. $^*$ indicates statistical 
significance ($p<0.001$) compared to the previous best result in the 
Overall column; $^{\mathrm{ns}}$ indicates no significant difference 
from the previous best ($p \geq 0.05$). CAHAL (T1w init.) denotes the 
framework initialised with T1-weighted pretrained weights.}
\label{tab:flair_psnr_crossmod}
\resizebox{\columnwidth}{!}{%
\begin{tabular}{lcccccccc}
\toprule
\textbf{Method} & \textbf{1.0--1.5} & \textbf{1.5--2.5} & \textbf{2.5--3.5} 
    & \textbf{3.5--4.5} & \textbf{4.5--5.5} & \textbf{5.5--6.5} 
    & \textbf{6.5--10.5} & \textbf{Overall} \\
 & ($\mathrm{mm}^3$) & ($\mathrm{mm}^3$) & ($\mathrm{mm}^3$) 
 & ($\mathrm{mm}^3$) & ($\mathrm{mm}^3$) & ($\mathrm{mm}^3$) 
 & ($\mathrm{mm}^3$) & (Mean $\pm$ Std) \\
\midrule
Reference (Input)
    & $37.448 \pm 2.951$ & $34.624 \pm 1.677$ & $33.399 \pm 0.554$
    & $32.473 \pm 0.577$ & $31.877 \pm 0.738$ & $31.114 \pm 0.688$
    & $30.209 \pm 0.981$ & $32.799 \pm 2.445$ \\
\midrule
\multicolumn{9}{l}{\textit{State-of-the-Art (Generative)}} \\
SuperSynth \parencite{Liu2025AImaging}
    & $30.013 \pm 0.424$ & $29.727 \pm 0.192$ & $29.444 \pm 0.282$
    & $29.225 \pm 0.386$ & $29.080 \pm 0.468$ & $28.790 \pm 0.546$
    & $28.327 \pm 0.925$ & $29.190 \pm 0.723^*$ \\
\midrule
\multicolumn{9}{l}{\textit{Proposed Framework (Residual, native space)}} \\
\textbf{CAHAL}
    & $\mathbf{43.391 \pm 2.092}$ & $\mathbf{40.833 \pm 1.256}$ & $\mathbf{39.432 \pm 0.810}$
    & $\mathbf{37.891 \pm 1.407}$ & $\mathbf{36.916 \pm 1.253}$ & $\mathbf{36.138 \pm 1.493}$
    & $\mathbf{34.686 \pm 2.137}$ & $\mathbf{38.224 \pm 3.006}^*$ \\
CAHAL (T1w init.)
    & $39.221 \pm 3.552$ & $39.143 \pm 2.309$ & $38.432 \pm 1.625$
    & $37.250 \pm 2.542$ & $36.411 \pm 2.451$ & $35.577 \pm 2.365$
    & $33.983 \pm 2.609$ & $37.041 \pm 3.067^{\mathrm{ns}}$ \\
\bottomrule
\end{tabular}
}
\end{table}

\end{document}